\documentclass[sigconf]{acmart}

\AtBeginDocument{%
  }

\copyrightyear{2026}
\acmYear{2026}
\setcopyright{cc}
\setcctype{by}
\acmConference[WWW '26]{Proceedings of the ACM Web Conference 2026}{April 13--17, 2026}{Dubai, United Arab Emirates}
\acmBooktitle{Proceedings of the ACM Web Conference 2026 (WWW '26), April 13--17, 2026, Dubai, United Arab Emirates}
\acmPrice{}
\acmDOI{10.1145/3774904.3792107}
\acmISBN{979-8-4007-2307-0/2026/04}

\usepackage{hyperref}       
\usepackage{url}            
\usepackage{booktabs}       
\usepackage{amsfonts}       
\usepackage{nicefrac}       
\usepackage{microtype}      
\usepackage{xcolor}         
\usepackage{amsmath}
\usepackage{lipsum}
\usepackage{graphicx}
\usepackage{multirow}
\usepackage{tabularx}
\graphicspath{{./figures/}{./figures/loss_landscape/}}
\usepackage[linesnumbered,ruled]{algorithm2e}
\usepackage{subcaption}
\usepackage{float}

\newcommand\our{GRN}
\newcommand\ours{GRNs}

\title{Graph Retention Networks for Dynamic Graphs}

\author{Qian Chang}
\affiliation{%
  \institution{University of Auckland}
  \city{Auckland}
  \country{New Zealand}
}
\affiliation{%
  \institution{Central China Normal University}
  \city{Wuhan}
  \country{China}
}
\email{qcha783@aucklanduni.ac.nz}

\author{Xia Li}
\authornote{Corresponding author.}
\affiliation{%
  \institution{Central China Normal University}
  \city{Wuhan}
  \country{China}
}
\email{lixia@ccnu.edu.cn}

\author{Xiufeng Cheng}
\affiliation{%
  \institution{Central China Normal University}
  \city{Wuhan}
  \country{China}
}
\email{xiufengcheng@ccnu.edu.cn}

\author{Runsong Jia}
\affiliation{%
  \institution{University of Technology Sydney}
  \city{Sydney}
  \country{Australia}
}
\email{runsong.jia@student.uts.edu.au}

\author{Jinqing Yang}
\affiliation{%
  \institution{Central China Normal University}
  \city{Wuhan}
  \country{China}
}
\email{yjq@mails.ccnu.edu.cn}

\author{Guoping Hu}
\affiliation{%
  \institution{University of Auckland}
  \city{Auckland}
  \country{New Zealand}
}
\email{ghu206@aucklanduni.ac.nz}

\author{Ciprian Doru Giurcaneanu}
\affiliation{%
  \institution{University of Auckland}
  \city{Auckland}
  \country{New Zealand}
}
\email{c.giurcaneanu@auckland.ac.nz}

\renewcommand{\shortauthors}{Qian Chang et al.}


\begin{document}

\begin{abstract}
In this paper, we propose \textbf{Graph Retention Networks} (\ours{}) as a unified architecture for deep learning on dynamic graphs. The \our{} extends the concept of retention into dynamic graph data as graph retention, equipping the model with three key computational paradigms: parallelizable training, low-cost $\mathcal{O}(1)$ inference, and long-term chunkwise training. This architecture achieves an optimal balance between \textit{efficiency}, \textit{effectiveness}, and \textit{scalability}. Extensive experiments on benchmark datasets demonstrate its strong performance in both edge-level prediction and node-level classification tasks with significantly reduced training latency, lower GPU memory overhead, and improved inference throughput by up to 86.7x compared to SOTA baselines. The proposed \our{} architecture achieves competitive performance across diverse dynamic graph benchmarks, demonstrating its adaptability to a wide range of tasks.
\end{abstract}
\maketitle

\section{Introduction}
\label{section-1}
Dynamic graphs offer a flexible and powerful geometric framework for modeling complex real-world systems that evolve over time, such as ecosystems \cite{li2021ecosystem,ntekouli2024exploiting}, social networks \cite{min2021social,alvarez2021evolutionary,kumar2019jodie}, and traffic systems \cite{yu2017spatio,wu2019graph,guo2021traffic}. These structures capture the temporal dynamics of interactions between entities, enabling a deeper understanding of processes that unfold across both topological and temporal dimensions. Dynamic graphs exhibit progressive growth in scale and connectivity as interactions evolve, increasing both the modeling complexity and the computational demands of learning from such structures. Research has increasingly focused on modeling temporal dependencies and evolving relationships \cite{battaglia2016interaction, yin2024dynamic, li2024robust}. Unlike sequence-based retention models developed for natural language processing which assume linear and position-ordered token streams, dynamic graphs are characterized by asynchronous events and evolving topological dependencies that cannot be faithfully reduced to a single sequence~\cite{sun2023retentive,rossi2020tgn}. This structural discrepancy motivates a graph-native retentive formulation that jointly models temporal memory and structural propagation, rather than adapting retention mechanisms originally designed for NLP settings. However, three main challenges hinder the \textit{efficiency}, \textit{effectiveness}, and \textit{scalability} of existing approaches as the prominence of dynamic systems grows.

\textit{One critical challenge leading to inefficiencies is the inability to reconcile training parallelism with low-cost inference} \cite{luo2024no,gao2024etc,wu2024feasibility,zhou2022tgl}, particularly for dense, large-scale, and long-term dynamic graphs. While large-scale dynamic graphs demand parallel computation to process ever-growing and increasingly dense temporal interactions, most existing models fail to maintain this efficiency for inference \cite{mondal2012managing,vatter2023evolution}. In practice, inference requires sequential or state-dependent retrieval of historical information, which breaks parallelism and introduces latency bottlenecks. This dual requirement imposes significant storage demands and computational overhead \cite{zhou2023disttgl,feng2024survey}, particularly in real-time inference scenarios where timely access to both historical and current state information is essential.

\textit{Another pivotal barrier to the effectiveness of dynamic graph learning lies in establishing a principled balance between long-term and short-term temporal dependencies} \cite{zhang2024towards,yu2023dygformer,rossi2020tgn,miller2024survey, vatter2023evolution, zhang2023llm4dyg}. Real-world dynamic systems often exhibit temporal heterogeneity, where certain relational patterns emerge gradually over extended durations, while others evolve rapidly within narrow temporal windows. Consequently, Temporal Graph Neural Networks (TGNNs) are expected to preserve long-term contextual information while remaining sensitive to recent fluctuations \cite{yu2023dygformer, huang2024tgb,pareja2020evolvegcn,xu2020tgat}. Many existing methods lack adaptive mechanisms to regulate this temporal trade-off, resulting in representations that are either overly reactive to recent inputs or overly biased toward stale historical signals.

\textit{In terms of scalability, most existing models rely on auxiliary modules to manage temporal information flow.} Typical approaches adopt separate components such as sampling or truncation strategies \cite{xu2020tgat,wang2021cawn,yu2023dygformer} and dedicated memory modules \cite{kumar2019jodie,weston2014memory,zhou2022tgl,su2024pres} to approximate historical context. While such designs can reduce computational overhead or increase temporal coverage, they inevitably introduce architectural fragmentation. Sampling and truncation strategies risk discarding salient temporal signals due to local heuristics or fixed temporal windows, whereas memory-based approaches suffer from scalability bottlenecks \cite{barros2021survey}. The absence of a unified graph operator exacerbates these issues, limiting the scalability and broader applicability across diverse dynamic graph scenarios.

\textbf{Contributions:}
This work proposes a unified and scalable architecture \ours{} that addresses the three key challenges in dynamic graph learning. (\romannumeral1) We propose a graph retention mechanism to enable efficient parallel training and low-cost inference with reduced latency, lower memory overhead, and improved computational throughput. (\romannumeral2) We introduce a multi-scale decay conjugated to a recurrent state that enables the model to balance long-term and short-term time dependencies for the representation improvement of various dynamic graph scenarios. (\romannumeral3) By consolidating memory and aggregation into a unified architectural design without reliance on truncation or sampling-based heuristics, the \our{} effectively eliminates modular inefficiencies and achieves strong scalability, generalizability, and architectural coherence across both continuous-time and discrete-time dynamic graphs.

\section{Related Work}
\label{section-2}
Early methods for learning on dynamic graphs typically utilized separate modules, with graph neural networks capturing structural information and recurrent neural networks handling temporal dependencies \cite{pareja2020evolvegcn,kumar2019jodie,manessi2020dgcn,xu2020tgat,rossi2020tgn}. While this modular approach was intuitive, it faced significant scalability challenges \cite{bonner2019tna,bai2022drain,li2023spikenet}, particularly when applied to large and densely connected graphs with extensive temporal information. Maintaining both structural and temporal embeddings significantly increased computational complexity \cite{cui2018survey,zhu2018proximity}, severely impacting training efficiency.

To enhance the scalability of TGNNs, researchers have adopted techniques such as neighbor truncation \cite{yu2023dygformer,manessi2020dgcn} and selective sampling \cite{zhang2024towards,rossi2020tgn,guan2022dynagraph,trivedi2019dyrep}. These approaches allow models to allocate computational resources more effectively by focusing on the most relevant temporal interactions. Such techniques are particularly advantageous for real-time reasoning, where rapid inference and the ability to model both short-term and long-term dependencies are essential. Recent advances in TGNNs have introduced strategies to improve training efficiency on large-scale dynamic graphs. These include advanced data batching techniques, optimized memory management pipelines, and GPU utilization strategies \cite{yu2023dygformer,wan2023g3}. By reducing data access costs \cite{gao2024etc}, minimizing redundant data handling \cite{guan2022dynagraph}, and streamlining data transfers \cite{mondal2012managing,qiu2018real,besta2019practice}, these innovations significantly boost throughput and lower computational burdens, enabling models to handle high-dimensional datasets effectively while maintaining their performance.

Unified architectures have emerged as a promising alternative, integrating topological and temporal information into a single framework. These models leverage unique graph operators \cite{zhou2022tgl,zhou2023ugrapher}, truncation strategies \cite{yu2023dygformer}, and temporal encoding functions \cite{huang2024tgb,cong2023graphmixer} to maintain temporal context without the need for post-training memory updates. Attention-based methods and causal masking further enhance these architectures by selectively aggregating information, enabling efficient capture of long-term dependencies in dynamic graph data while avoiding computational inefficiencies.
\section{Preliminaries}
\label{section-3}


\subsection{Dynamic Graphs}
\label{section-3:dynamicgraphs}
A graph $\mathcal{G} = (\mathcal{V}, \mathcal{E})$ is an irregular data structure composed of a set of nodes (also referred to as vertices) $\mathcal{V} = \{v_1, v_2, \dots, v_{\lvert{\mathcal{V}}\rvert}\}$ and edges $\mathcal{E} = \{(v_i,v_j) \mid i, j \in \left\{ 1, 2, \dots, {\lvert{\mathcal{V}}\rvert}\right\}\}$, where each edge $\left( v_i,v_j\right)$ represents a connection between nodes $v_i$ and $v_j$. Both the nodes and edges can be associated with features, denoted by $X = \{x_1, x_2, \dots, x_{\lvert{\mathcal{V}}\rvert}\}$ and $E = \{e_{ij}|(v_i,v_j)\in \mathcal{E}\}$, respectively, which encode information relevant to the entities and structures within the graph. In dynamic graphs, temporal attributes are introduced to capture changes in both node and edge structures over time. These graphs evolve either continuously or discretely as nodes and edges appear, disappear, or update their properties over time. Dynamic graphs are generally categorized into two main types due to their inherent temporal properties.
\textbf{Discrete-Time Dynamic Graphs} (DTDGs) evolve in distinct time-ordered snapshots $\left\{\mathcal{G}_1, \mathcal{G}_2, \dots, \mathcal{G}_T\right\}$, where each graph $\mathcal{G}_t = (\mathcal{V}_t, \mathcal{E}_t)$ represents the state of the graph at a specific time-step $t$. Temporal updates occur at discrete intervals, and each snapshot captures the state of the nodes and edges at that particular time. This formulation is particularly well-suited for applications where data are collected or processed at regular intervals.
In contrast, \textbf{Continuous-Time Dynamic Graphs} (CTDGs) allow changes to occur continuously over time, with nodes and edges evolving irregularly, and events being associated with specific timestamps. The graph is represented as a function of continuous time $\mathcal{G}(t) = (\mathcal{V}(t), \mathcal{E}(t))$, where $t$ is the timestamp. In this framework, nodes and edges can be updated or interact at any point in time, making CTDGs particularly suitable for modeling asynchronous events.

\subsection{Temporal Graph Neural Networks}
\label{subsec:background:TGNN}
We follow the standard Temporal Graph Neural Networks (TGNN) paradigm to describe how the \our{} works. A typical TGNN learns representations using a dynamic message-passing paradigm, which aggregates information from the neighborhood of each node over time:
\begin{equation}
\begin{aligned}
\label{eq:tgnn}
\underbrace{v_{i}^{(k)}\gets \gamma }_{\text{update}} \left (v_{i}^{(k-1)}, \underbrace{ \underset{{j \in \mathcal{N }(i) }}{\oplus} \overbrace{ \phi \left ( v_{i}^{(k-1)}, v_{j}^{(k-1)}, e_{ij},t\right )}^{\text{message} } }_{\text{aggregate}}\right ) 
\end{aligned}
\end{equation}
where $\phi(\cdot)$ is the \textbf{message function} that generates messages between nodes $v_i$ and $v_j$ using their features and edge features $e_{ij}$ at time $t$, $\oplus$ denotes the \textbf{aggregation function} that combines messages from all neighboring nodes $j\in\mathcal{N}(i)$, and $\gamma(\cdot)$ is the \textbf{update function} that integrates the aggregated messages and the previous state of the node to update its representation over time. This process produces node representations that are subsequently utilized for various downstream tasks.

In this study, we evaluate the efficiency, effectiveness, and scalability of representation learning through two typical tasks in dynamic graph learning: (i) \textbf{edge-level prediction}, which predicts whether $v_i$ and $v_j$ are connected at time $t$, and (ii) \textbf{node-level classification}, which infers the state label of $v_i$ at time $t$.
\section{Graph Retention Networks}
\label{section-4}
\label{sec:\our{}}
Graph Retention Networks (\ours{}) consist of a stack of multiple \our{} blocks, each integrating a multi-scale graph retention module. Within each block, the core computational operator is the graph retention mechanism, which selectively retains and propagates crucial graph information across temporal updates. This retention operator is paired with a feedforward network that refines and transforms the retained graph representations, enabling effective performance on downstream tasks.

\subsection{Graph Retention}

Graph retention operates as a dual-form graph operator that aggregates information from source and destination nodes in either a parallel or recurrent manner. As described in Section~\ref{section-3:dynamicgraphs}, consider a destination node \(v_t^i\) at time step \(t\), represented by the feature vector \(\mathbf{x}_t^i \in \mathbb{R}^d\), and its corresponding set of source nodes (i.e., historical 1-hop neighbors) \(\{ v_t^j \mid j \in \mathcal{N}(i) \}\).
The embeddings of the source nodes \(\{ \mathbf{x}_t^j \in \mathbb{R}^d \mid j \in \mathcal{N}(i) \}\) and their associated edge attributes \(\{ \mathbf{e}_t^j \in \mathbb{R}^d \mid j \in \mathcal{N}(i) \}\), ordered by interaction time, are organized into matrices \(\mathbf{X}^j \in \mathbb{R}^{|\mathcal{N}(i)| \times d}\) and \(\mathbf{E}^j \in \mathbb{R}^{|\mathcal{N}(i)| \times d}\), respectively.
The time intervals \(\{ \Delta t_j \mid j \in \mathcal{N}(i) \}\), representing the elapsed time between each interaction and the current time, are collected into a temporal offset vector denoted as \(\Delta \mathbf{T}\).

The \textbf{message function} is then applied to map the destination node features, edge attributes, and time intervals into a new feature space through a simple linear transformation, denoted as $\phi(\cdot)$. We compute the transformed representation via 
$\mathbf{X}^{j}=\phi\left(\mathbf{X}^{j}, \mathbf{E}^{j},\Delta{\mathbf{T}} \right)=\mathbf{X}^{j}+\mathbf{E}^{j}\mathbf{W}_{e}+\mathrm{TE}\left(\Delta \mathbf{T} \right)$
where $\mathbf{W}_{e}^{j}\in \mathbb{R}^{{d} \times d}$ is a learnable weight matrix, and $\mathrm{TE}(\cdot)$ represents the temporal encoding function as described in Section \ref{sec:TE}.

\begin{figure*}[!t]
    \centering
    \includegraphics[width=0.85\textwidth]{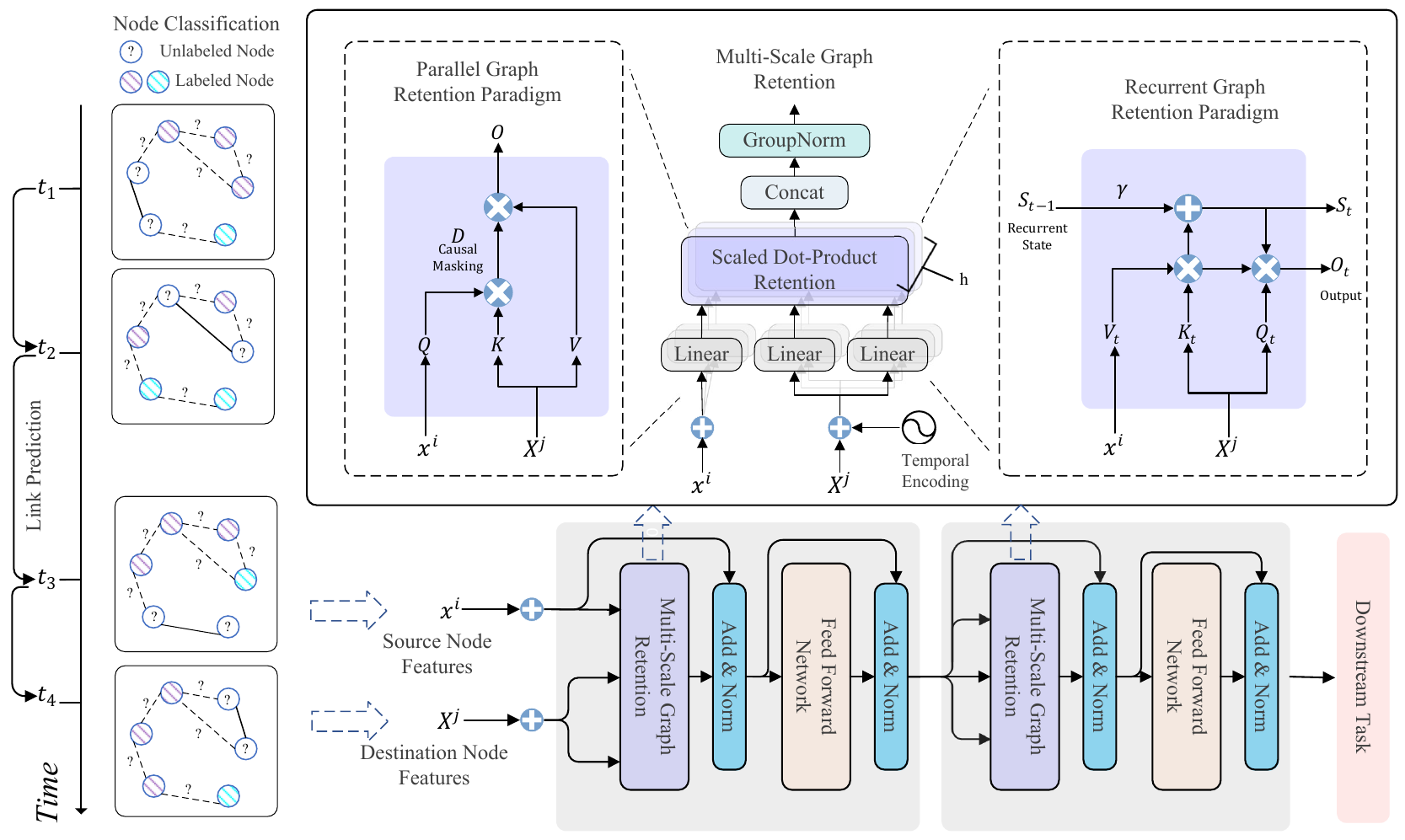}
    \label{fig:arch}
    \caption{Architecture of \ours{}.}
\end{figure*}

Next, we denote graph retention as an \textbf{aggregation function} $\oplus \left( \mathbf{x}_{t}^{i}, \mathbf{X}^{j}\right)$ that maps $\mathbf{x}_{t}^{i}\stackrel{agg}{\longmapsto}{o}_{t}^{i}$. We begin by projecting $\mathbf{x}_{i}$ onto a query vector and $ \mathbf{X}^{j}$ onto the key-value pairs as follows:
\begin{equation}
\mathbf{q} = \mathbf{x}^{i}\mathbf{W}_{q} + \mathbf{b}_{q}, \quad
\mathbf{K} = \mathbf{X}^{j}\mathbf{W}_{k} + \mathbf{1}^{\top}\mathbf{b}_{k}, \quad
\mathbf{V} = \mathbf{X}^{j}\mathbf{W}_{v} + \mathbf{1}^{\top}\mathbf{b}_{v},
\label{eq:qkv}
\end{equation}
where
$\mathbf{W}_q, \mathbf{W}_k, \mathbf{W}_v \in \mathbb{R}^{d \times d}$
are learnable weight matrices,
$\mathbf{b}_q, \mathbf{b}_k, \mathbf{b}_v \in \mathbb{R}^{d}$
are bias vectors,
and $\mathbf{1}^{\top}$ denotes an

all-ones column vector of $ \mathbb{R}^{|\mathcal{N}(i)| \times 1}$.

Following the idea of retention mechanism computation \cite{sun2023retentive}, we maintain a recurrent state $\mathbf{S}^{i}$ for the source node $v_{t}^{i}$. The output of the graph retention mechanism is then computed in a recurrent manner:
\begin{equation}
\begin{aligned}
\label{eq:recurrent}
&\mathbf{S}_{t}^{i}=\lambda \mathbf{S}_{t-1}^{i}+\mathbf{K}_{t}^{\intercal}\mathbf{V}_{t},
\hfill
& \mathbf{K}_{t}\in\mathbb{R}^{1 \times d},
\mathbf{V}_{t}\in\mathbb{R}^{1 \times d}
 \\
&\mathbf{o}_{t}^{i}= w_{t} \mathbf{q}_{t} \mathbf{S}_{t}^{i}=  w_{t}\mathbf{q}_{t} \sum_{k=1}^{t} \lambda^{t-k} \mathbf{K}_{k}^{\intercal}\mathbf{V}_{k},
\hfill & \mathbf{q}_{t}\in\mathbb{R}^{1\times d}
\end{aligned}
\end{equation}
where the source node is aggregated with the destination node using the state tensor $\mathbf{S}_{t}^{i}$, $w_t$ denotes the edge weight applied at time step $t$, and $\lambda\in(0,1)$ is the temporal decay factor to reduce the interference of remote information to current information over time steps. The graph retention formulation is inherently parallelizable within training instances. To accommodate different training and inference scenarios, we reformulate Equations (\ref{eq:qkv}) and (\ref{eq:recurrent}) as:

\textbf{Parallel Paradigm of Graph Retention.}
Unlike sequence modeling, node-wise modeling in dynamic graphs excludes self-interactions within the neighboring node set, but rather cross-attention between the source node and its set of neighboring nodes. The parallel graph retention paradigm is formulated as follows:
\begin{equation}
\begin{aligned}
\mathbf{Q}=\mathbf{X}^{i}\mathbf{W}_{q}+\mathbf{1}^{\top}\mathbf{b}_{q},
\quad
\mathbf{K} & =\mathbf{X}^{j}\mathbf{W}_{k}+\mathbf{1}^{\top}\mathbf{b}_{k},
\quad
\mathbf{V}=\mathbf{X}^{j}\mathbf{W}_{v}+\mathbf{1}^{\top}\mathbf{b}_{v}\\
\mathbf{D}_{tk} & =
\left\{
\begin{aligned}
& \lambda^{t-k} w_{t}, && t \ge k \\
& 0, && t<k \\
\end{aligned}
\right.
\\
\mathrm{GraphRete}& \mathrm{ntion}(\mathbf{x}^{i},\mathbf{X}^{j}) = ( \mathbf{Q} \mathbf{K}^\intercal \odot \mathbf{D})\mathbf{V}
\end{aligned} 
\end{equation}
where $\mathbf{D}$ is a weighted causal mask, with indices $t,k \in \{1,\ldots,|\mathcal{N}(i)|\}$ corresponding to the temporal order of source nodes in the neighborhood of $v^i$. This mask ensures that future source nodes remain invisible to the current target node, thereby preserving the causal structure during computation. The symbol $\odot$ denotes the Hadamard product.

\textbf{Recurrent Paradigm of Graph Retention.}
The recurrent graph retention paradigm enables the model to iteratively aggregate information across sequential time steps, making it well-suited for long-term inference in dynamic graphs. At each time-step $t$, the model updates the recurrent state $\mathbf{S}$ based on the current key-value interaction: 
\begin{equation}
\begin{aligned}
\label{eq:rmanner}
&\mathbf{S}_{t}=\lambda \mathbf{S}_{t-1}+\mathbf{K}_{t}^{\intercal}\mathbf{V}_{t}
\hfill \\
&\mathrm{GraphRetention} (\mathbf{x}_{t}^{i},\mathbf{X}^{j}) = w_{t}\mathbf{q}_t \mathbf{S}_t, \quad t = 1, \cdots, {|\mathcal{N}(i)|} \\
\end{aligned}
\end{equation}
This recurrent mechanism enables \ours{} to retain and update node embeddings over time at a low computational cost, thereby facilitating the effective learning of complex temporal dependencies.

\textbf{Chunkwise Paradigm of Graph Retention.}
To enable efficient training on long-term and dense CTDGs, we utilized a hybrid computational paradigm combining parallel and recurrent approaches. This paradigm splits CTDGs into multiple stages, treating each stage as a distinct training batch, thereby conserving memory by training the model in manageable chunks. The chunkwise graph retention paradigm for the $m$-th stage is formulated as follows:
\begin{equation}
\begin{aligned}
\mathbf{Q}_{[m]}=&\mathbf{X}_{Bm:B(m+1)}^{i}\mathbf{W}_{q}+\mathbf{1}^{\top}\mathbf{b}_{q},
\\
\mathbf{K}_{[m]}=&\mathbf{X}_{Bm:B(m+1)}^{j}\mathbf{W}_{k}+\mathbf{1}^{\top}\mathbf{b}_{k},
\\
\mathbf{V}_{[m]}=&\mathbf{X}_{Bm:B(m+1)}^{j}\mathbf{W}_{v}+\mathbf{1}^{\top}\mathbf{b}_{v}
\\
\mathbf{R}_{[m]} =& \lambda^{B}\mathbf{R}_{[m-1]} + \mathbf{K}_{[m]}^\intercal \mathbf{V}_{[m]}
\\
\mathrm{Graph}\mathrm{Retention}(\mathbf{x}^{i},\mathbf{X}^{j})& = 
\underbrace{ (\mathbf{Q}_{[m]} \mathbf{K}_{[m]}^\intercal \odot D)\mathbf{V}_{[m]}}_\text{current chunk}
 + \underbrace{\lambda^{B} \mathbf{Q}_{[m]}\mathbf{R}_{[m-1]}}_\text{past chunk}
\end{aligned}
\end{equation}
where $B$ is the chunk size and $\mathbf{W}_q, \mathbf{W}_k, \mathbf{W}_v$ are identical to those defined in Equation (\ref{eq:qkv}). The chunkwise paradigm without causal masking can also be applied to DTDGs.

Finally, the output of the graph retention mechanism serves as the representation for downstream tasks. The destination node embedding is assigned the output representation via \textbf{update function} for the subsequent iteration as $\mathbf{x}_{t+1}^{i}={o}_{t}^{i}$. 

\subsection{Temporal Encoding}
\label{sec:TE}
In this section, we introduce a temporal encoding mechanism from GraphMixer \cite{cong2023graphmixer} into our architecture. This mechanism enables destination nodes to become time-aware of their neighboring nodes. Considering a specific time interval $\Delta t$, we define the time-encoding function as follows:
\begin{equation}
\begin{aligned}
\label{eq:time-encoding}
\mathrm{TE}\left(\Delta t,i\right) = \cos \left(\Delta t \cdot \sqrt{d}^{-(i - 1)/\sqrt{d}} \right)
\end{aligned}
\end{equation}
where $i$ is the dimension of the encoding space, and $d$ denotes the overall dimension of the encoding space. Temporal encoding not only enhances the richness of the feature space but also aligns the model's understanding with the intrinsic temporal characteristics of the underlying graph data.

\subsection{Multi-Scale Graph Retention}
Multi-Scale Graph Retention (MGR) extends the graph retention mechanism by enabling the model to focus simultaneously on information from multiple representation subspaces across different time intervals. This multi-head approach, inspired by \cite{vaswani2017transfomers}, captures complex temporal and relational dynamics, enabling the learning of robust representations in dynamic graphs.
In the MGR layer, we use $h$ heads, with each head independently performing graph retention.
\begin{equation}
\begin{aligned}
\label{eq:MGR}
\mathrm{head}_{i}&=\mathrm{GraphRetention}\left( \mathbf{X}^{i}, \mathbf{X}^{j} \right) \\
\mathrm{MGR}\left ( \mathbf{X}^{i}, \mathbf{X}^{j} \right ) &= \mathrm{GN} \left( \mathrm{Concat}\left( \mathrm{head}_{1},\mathrm{head} _{2},\dots,\mathrm{head} _{h}\right)\right)
\end{aligned}
\end{equation} 
where each head is associated with distinct decay $\lambda$ scales and projection parameters $\mathbf{W}_q, \mathbf{W}_k, \mathbf{W}_v \in\mathbb{R}^{d \times d} $. Group normalization (GN) \cite{wu2018group} is applied to the concatenated outputs of the heads to stabilize training and maintain numerical precision. This ensures consistency across heads, even when the scales and distributions of the information vary.

\textbf{Graph Retention Score Normalization.}
The \our{} requires normalization of the graph retention scores to maintain numerical stability within the computational process, i.e., (\romannumeral1) We normalize $\mathbf{QK}^\top$ as ${\mathbf{QK}^\top}/{\sqrt{d}}$; (\romannumeral2) We replace $\mathbf{D}$ with a scaled version $\tilde{\mathbf{D}}_{nm}$, where $\tilde{\mathbf{D}}_{nm} = {\mathbf{D}_{nm}}/{\sum_{i=1}^{n} \mathbf{D}_{ni}}$; (\romannumeral3) Let $\mathbf{R}$ denote the retention scores, where $\mathbf{R} = \mathbf{QK}^\top \odot \mathbf{D}$, we further normalize $\mathbf{R}$ as $\tilde{\mathbf{R}}_{nm} = {\mathbf{R}_{nm}}/{\max\left( \left| \sum_{i=1}^{n} \mathbf{R}_{ni} \right|, 1 \right)}$. These operations do not influence the output of graph retention, owing to the GroupNorm.

\subsection{Overall Architecture of Graph Retention Networks}

Each \our{} block contains an MGR layer and a nodewise feedforward network (FFN) module, which together form the core components of the architecture. For an $N$-layer GRN, $N$ \our{} blocks are stacked to construct a deep-learning network capable of capturing complex dependencies and structural nuances in dynamic graphs. The FFN module consists of two linear transformations, with an h-Swish \cite{howard2019hswish} activation gate applied between them to introduce nonlinearity. We provide the overall process of \ours{} in Algorithm \ref{algorithm}.

For simplicity, we denote the inputs $\left\{\mathbf{X}^{i}, \mathbf{X}^{j} \right\}$ as $X$, and the embedding dimension of the FFN is defined as $d_{model}=h \times d$, where $h$ is the number of heads and $d$ is the dimension of each head. The general framework of \ours{} is defined as follows:
\begin{equation}
\begin{aligned}
\label{eq:archi}
H &= \mathrm{MGR}\left ( \mathrm{LN} \left ( X \right ) \right ) + X\\
O &= \mathrm{FFN}\left ( \mathrm{LN} \left ( H \right ) \right ) + H\\
\text{where } &\mathrm{FFN} \left ( x \right ) = \left ( \mathrm{hSwish} \left ( xW_1 \right ) \right ) W_2
\end{aligned}
\end{equation}
where $\mathrm{hSwish} (x)=x\cdot \mathrm{ReLU6} (x+3)/6=x \cdot \mathrm{min} \left ( \mathrm{max} \left(0,x+3 \right ),6\right)/6$, $\mathrm{LN} \left ( \cdot \right )$ represents layer normalization \cite{ba2016layernorm}, and $W_{1},W_{2}$ are learnable weight matrices in the FFN used for integrating the multiple retention outputs generated by the MGR.
\section{Experiments}
\label{section-5}

\begin{figure*}[!thb]
\centering
\resizebox{0.95\textwidth}{!}{
\begin{minipage}{\textwidth}
    \begin{subfigure}{1\textwidth}
    \centering
    \includegraphics[width=0.9\textwidth]{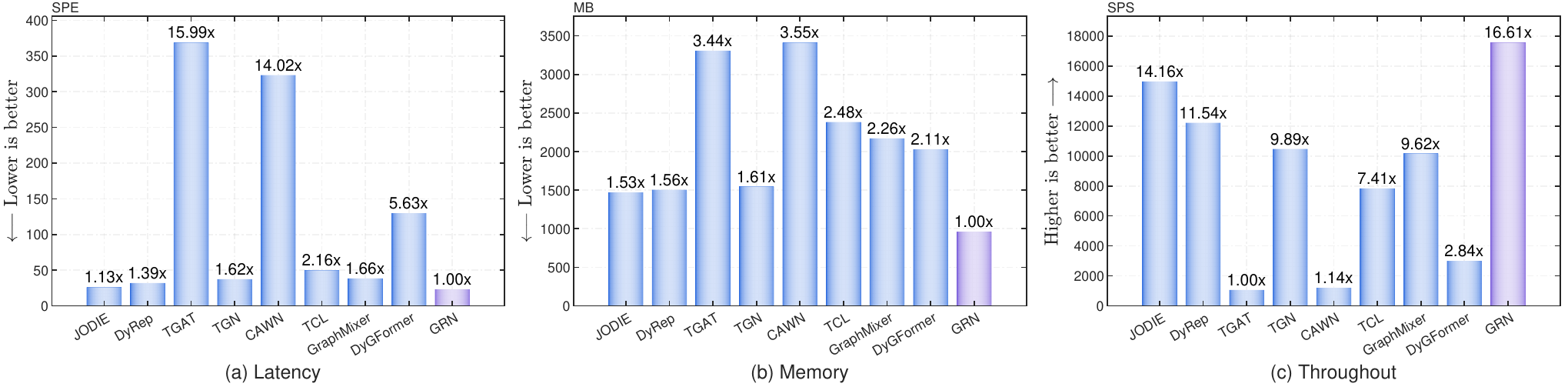}
    \caption{Training performance.}
    \label{fig:perf:train}
    \end{subfigure}
    \hfill
    \begin{subfigure}{1\textwidth}
    \centering
    \includegraphics[width=0.9\textwidth]{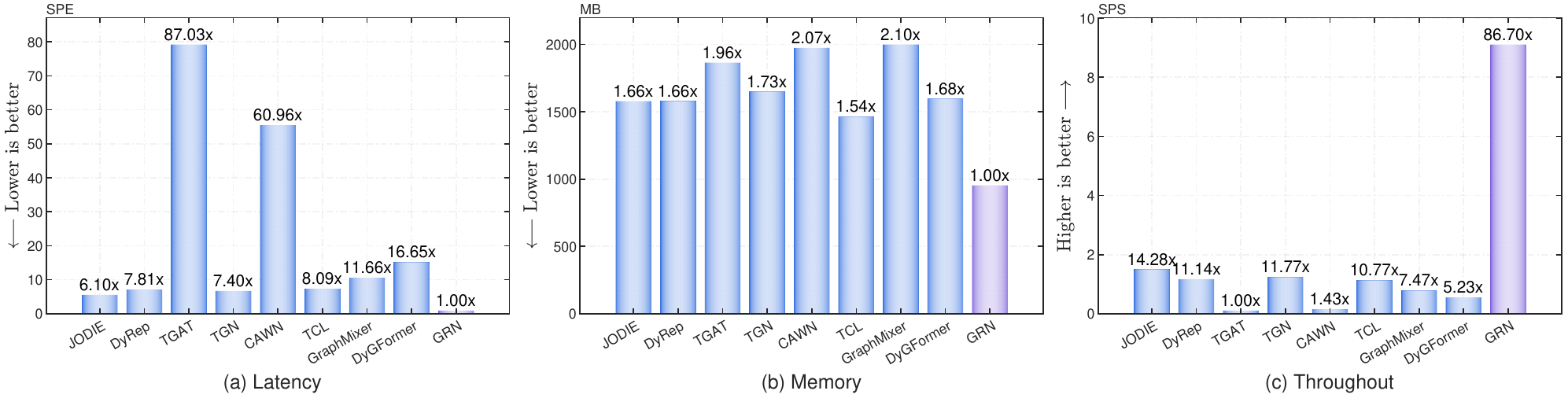}
    \caption{Inference performance.}
    \label{fig:perf:inference}
    \end{subfigure}
    \caption{Training and inference efficiency comparison. SPE = Seconds Per Epoch; MB = MegaByte; SPS = Samples Per Second. Subfigure (a) compares performance of baseline with that of \ours{} when trained in chunkwise paradigm, and subfigure (b) compares performance of \ours{} inference in recurrent paradigm with that of baselines. All multipliers are based on model with the lowest value.}
    \label{fig:perf:both}
\end{minipage}
}
\end{figure*}

\subsection{Setup}
\label{section-5:setup}

\textbf{Datasets.}
We conduct experiments on thirteen benchmark datasets (i.e., Wikipedia, Reddit, MOOC, LastFM, Myket, Enron, UCI, Flights, Can. Parl., US Legis., UN Trade, UN Vote, and Contact) from various domains, as collected by \cite{poursafaei2022towards}. The statistics of the datasets are summarized in Table~\ref{tab:datasets}, with more details provided in Appendix~\ref{appendix:dataset}.

\textbf{Baselines.}
We compare the \our{} with nine strong baselines: JODIE \cite{kumar2019jodie}, DyRep \cite{trivedi2019dyrep}, TGAT \cite{xu2020tgat}, TGN \cite{rossi2020tgn}, CAWN \cite{wang2021cawn}, EdgeBank \cite{poursafaei2022towards}, TCL \cite{wang2021tcl}, GraphMixer \cite{cong2023graphmixer}, and DyGFormer \cite{yu2023dygformer}. These baselines encompass a variety of architectures, including GNNs, memory-based methods, transformers, and random walk approaches. While these models utilize truncation and sampling strategies to aggregate past information, our model retains a recurrent state for graph retention. The sophisticated designs of these baselines reflect current advancements in improving representation learning for dynamic graphs. Detailed descriptions of the baselines are provided in Appendix \ref{appendix:baseline}.

\textbf{Downstream Tasks and Evaluation Metrics.}
We evaluate our model on edge-level prediction and node-level classification tasks. Specifically, we integrate our model with a link predictor and node classifier as downstream components. These components consist of multiple fully connected layers, with the link predictor estimating the probability of a connection between two nodes and the node classifier predicting the probability of a node belonging to a specific state label. In addition, we evaluate the edge-level prediction task under both transductive and inductive settings. In dynamic graphs, transductive learning involves observing node and edge information during training, allowing the model to make predictions based on a known structure. In contrast, inductive learning requires the model to predict links between nodes that remain unobserved during training. For performance evaluation, we use Average Precision (AP) to assess model precision and the Area Under the Receiver Operating Characteristic Curve (AUC-ROC) to evaluate model sensitivity and robustness for both prediction and classification tasks. To measure training and inference efficiency, we compared models based on training and inference latency, GPU memory consumption, and throughput.

\textbf{Implementation and Configurations.}
We follow the standard training pipeline and evaluation protocol proposed by the Dynamic Graph Library (DyGLib)\footnote{\url{https://github.com/yule-BUAA/DyGLib}} and the Temporal Graph Benchmark (TGB)\footnote{\url{https://github.com/shenyangHuang/TGB}} to implement our model to ensure consistency of comparisons. Nevertheless, we clarify in Appendix \ref{appendix:implementationclarification} that our implementation has some inconsistencies with it. The chunk-wise paradigm is used for training, while the recurrent computation paradigm is adopted for fast inference to ensure the retention of identical long-term dependencies. For baseline models, we use the optimal configurations reported by DyGLib after validation, while the optimal configurations for our model and computer resources for implementation are provided in Appendix \ref{appendix:configurations}. The implementation code is available at \href{https://anonymous.4open.science/r/GraphRetentionNet}{[Code]}.

\begin{table*}[!thb]
\centering
\resizebox{0.95\textwidth}{!}{
\begin{minipage}{1.1\textwidth}
\centering
\caption{AP for transductive link prediction on dynamic graphs with random negative sampling strategy. The best performances are highlighted in bold; second-best results are underlined.}
\label{tab:transAP}
\begin{tabular}{l|cccccccccc}
\hline
\toprule
Datasets & JODIE & DyRep & TGAT & TGN & CAWN & EdgeBank & TCL & GraphMixer & DyGFormer & GRN \\
\midrule
Wikipedia & 91.22±3.73 & 90.94±0.48 & 94.00±0.31 & 97.50±0.06 & \underline{98.41±0.02} & 90.37±0.00 & 95.18±0.14 & 95.91±0.05 & \textbf{98.86±0.04} & 97.49±0.35 \\ 
Reddit & 97.61±0.17 & 97.44±0.12 & 96.64±0.04 & 96.96±0.16 & 98.79±0.02 & 94.71±0.00 & 93.25±0.18 & 95.41±0.04 & \underline{98.98±0.06} & \textbf{99.50±0.03} \\ 
MOOC & 75.66±5.05 & 68.74±1.33 & \underline{77.73±1.23} & 74.73±1.65 & 68.42±1.25 & 57.97±0.00 & 75.19±1.34 & 74.23±0.40 & 72.12±1.16 & \textbf{91.63±1.18} \\ 
LastFM & 69.05±0.74 & 69.23±1.69 & 64.05±0.40 & 56.13±2.82 & \underline{83.09±0.05} & 79.29±0.00 & 59.33±0.22 & 69.64±0.03 & \textbf{90.05±0.03} & 72.93±0.62 \\ 
Myket & \textbf{87.67±0.08} & \underline{87.49±0.09} & 65.54±5.14 & 83.04±1.81 & 85.40±0.19 & 57.31±0.00 & 68.91±2.66 & 86.55±0.01 & 79.91±0.46 & 82.94±2.51 \\ 
Enron & 78.38±1.24 & 71.71±3.02 & 67.09±0.53 & 74.41±1.56 & \underline{86.01±0.66} & 83.53±0.00 & 69.37±1.55 & 80.99±0.11 & \textbf{87.42±2.09} & 83.15±2.49 \\ 
UCI & 80.38±2.39 & 52.14±2.64 & 78.88±0.54 & 87.15±0.93 & \underline{92.88±0.21} & 76.20±0.00 & 85.13±1.02 & 91.91±0.70 & \textbf{95.29±0.05} & 80.55±1.38 \\ 
Flights & 92.38±0.64 & 90.22±0.78 & 90.96±0.03 & 92.67±1.60 & \underline{96.88±0.23} & 89.35±0.00 & 89.59±0.17 & 89.69±0.01 & \textbf{98.72±0.03} & 96.63±1.46 \\ 
Can. Parl. & 73.18±0.77 & 67.08±0.93 & 67.51±1.09 & 68.84±1.03 & 67.67±2.70 & 64.55±0.00 & 66.04±2.46 & 75.37±0.71 & \textbf{94.49±0.92} & \underline{91.91±0.23} \\ 
US Legis. & \underline{78.45±0.35} & 68.72±3.28 & 61.36±3.91 & 70.91±0.23 & 66.16±1.42 & 58.39±0.00 & 60.38±0.56 & 70.16±0.56 & 72.42±0.40 & \textbf{79.21±3.82} \\ 
UN Trade & \underline{65.26±0.67} & 63.67±1.13 & 59.97±0.78 & 60.90±0.66 & 61.76±0.15 & 60.41±0.00 & 60.20±0.33 & 54.19±4.33 & 57.04±1.30 & \textbf{71.54±0.68} \\ 
UN Vote & \underline{63.12±1.41} & 60.17±0.71 & 52.13±0.42 & 57.15±1.09 & 51.82±0.15 & 58.49±0.00 & 51.35±0.13 & 51.45±0.23 & 53.13±0.11 & \textbf{83.09±1.33} \\ 
Contact & 93.04±0.23 & 84.66±2.48 & \underline{93.42±0.17} & 91.92±2.05 & 87.26±0.20 & 92.58±0.00 & 89.01±0.50 & 89.28±0.31 & \textbf{97.89±0.05} & 89.34±0.75 \\ 
\midrule
Avg. Rank & 4.15 & 6.54 & 6.85 & 5.23 & 4.62 & 7.54 & 7.92 & 5.92 & \underline{3.15} & \textbf{3.08} \\
\bottomrule
\end{tabular}
\end{minipage}
}
\end{table*}

\subsection{Performance Comparison with Baselines}
\label{section-5:comparison}
This section gives the results of comparative experiments evaluating the \emph{effectiveness}, \emph{efficiency}, and \emph{scalability} of GRN, with further details provided in Appendix~\ref{appendix:Additional Experimental Results}. We report results (\romannumeral1) on both link prediction and node classification tasks and (\romannumeral2) under transductive and inductive settings. Additionally, we assess training and inference efficiency (\romannumeral3) by measuring latency, GPU memory consumption, and throughput.

\textbf{Efficiency.} \textit{The \our{} enjoys more efficient training and inference in dynamic graph representation learning.} We evaluate the efficiency of training and inference in Figure~\ref{fig:perf:both}. During training, the \our{} exhibits latency reductions ranging from 1.13x to 15.99x, GPU memory consumption reductions of 1.53x to 3.55x, and throughput improvements of 1.17x to 16.61x compared to baseline models. The differences in inference efficiency are even more pronounced, with the \our{} achieving at least a 6.1x reduction in latency, 1.66x lower GPU usage, and up to an 86.7x increase in throughput. These improvements are grounded in two architectural properties: (\romannumeral1) The \our{} adopts a unified state-based design that obviates the need for explicit memory modules and a single recurrent state tensor suffices to preserve long-term temporal dependencies. (\romannumeral2) The \our{} leverages a dual-paradigm execution strategy: the parallelism-enabled chunkwise mode significantly reduces training memory overhead, while the recurrent mode allows rapid inference with minimal synchronization. A formal analysis of time and space complexity of the \ours{} is provided in Appendix~\ref{appendix:complexity}.

\begin{figure*}[!th]
  \centering
  \begin{subfigure}[b]{0.248\textwidth}
    \includegraphics[width=\linewidth]{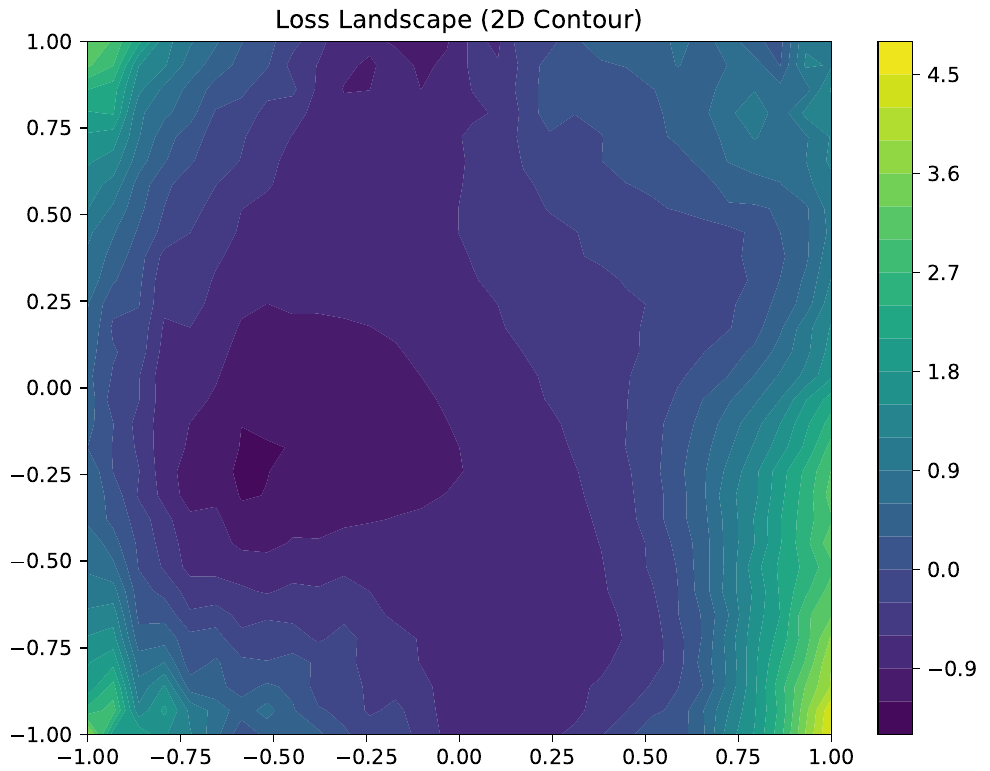}
    \subcaption{JODIE}
  \end{subfigure}%
  \begin{subfigure}[b]{0.248\textwidth}
    \includegraphics[width=\linewidth]{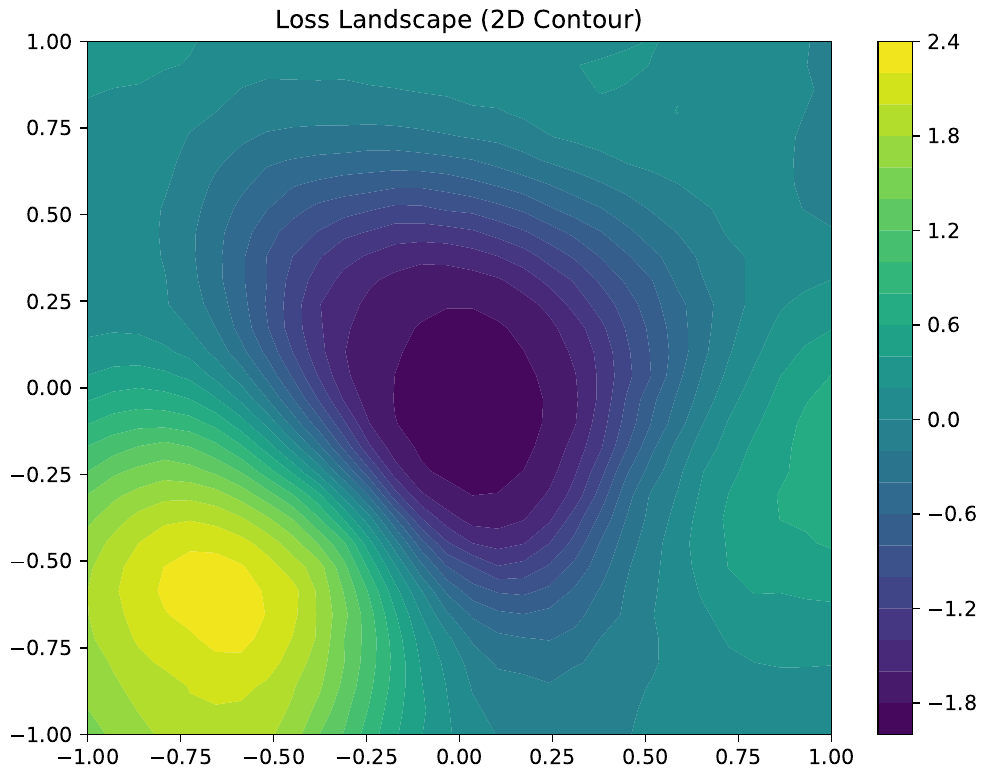}
    \subcaption{CAWN}
  \end{subfigure}%
  \begin{subfigure}[b]{0.248\textwidth}
    \includegraphics[width=\linewidth]{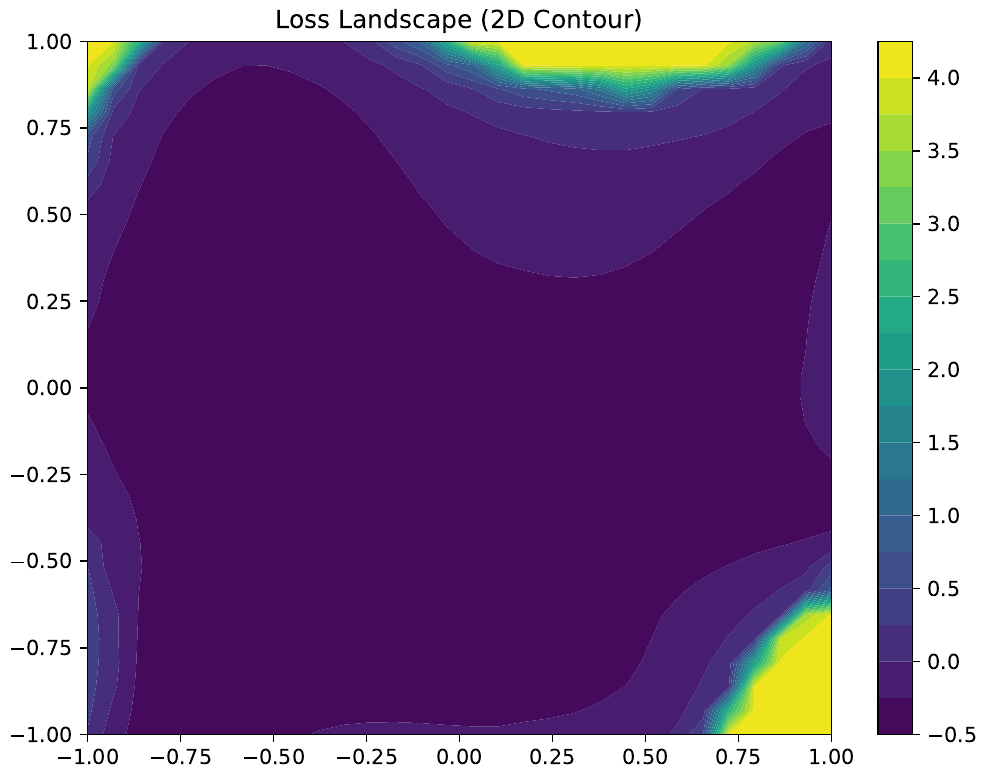}
    \subcaption{DyGFormer}
  \end{subfigure}%
  \begin{subfigure}[b]{0.248\textwidth}
    \includegraphics[width=\linewidth]{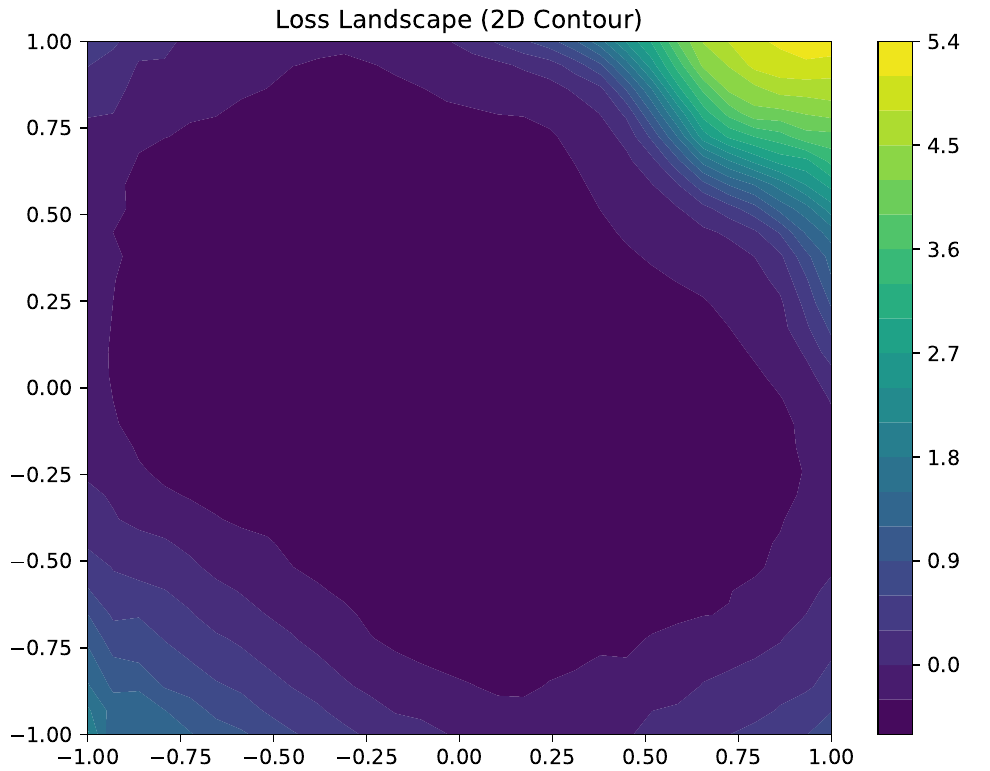}
    \subcaption{GRN}
  \end{subfigure}

  \begin{subfigure}[b]{0.248\textwidth}
    \includegraphics[width=\linewidth]{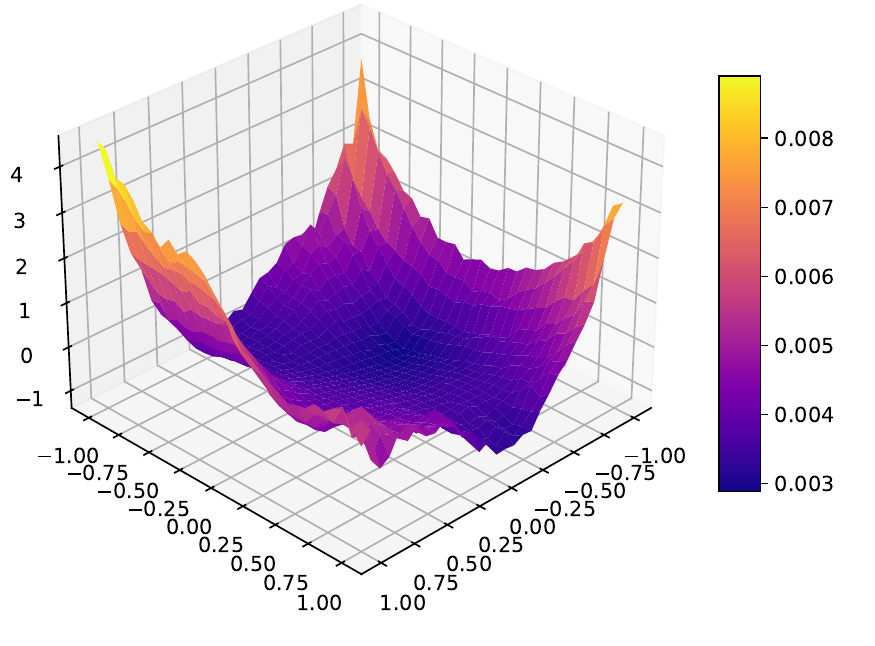}
    \subcaption{JODIE}
  \end{subfigure}%
  \begin{subfigure}[b]{0.248\textwidth}
    \includegraphics[width=\linewidth]{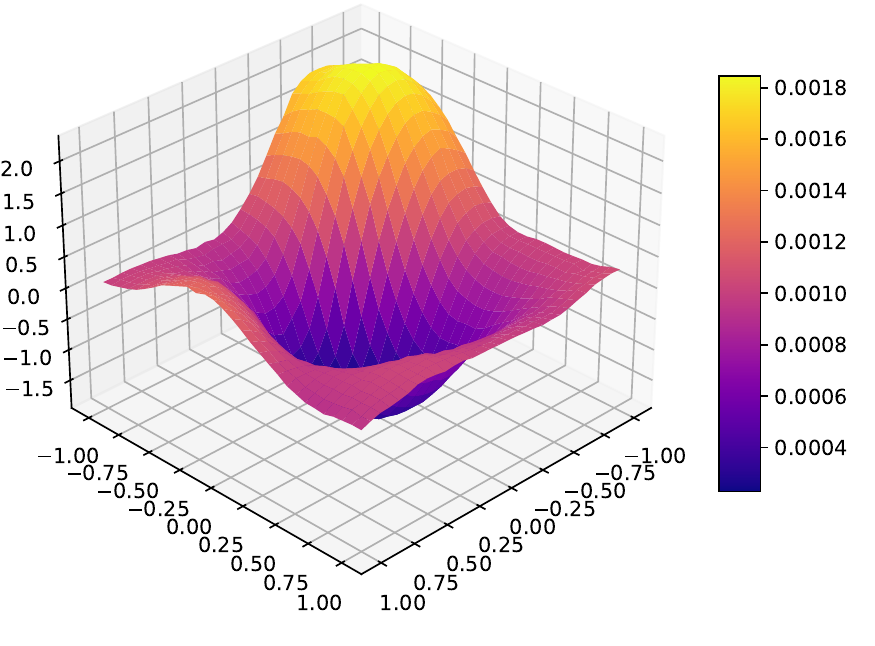}
    \subcaption{CAWN}
  \end{subfigure}%
  \begin{subfigure}[b]{0.248\textwidth}
    \includegraphics[width=\linewidth]{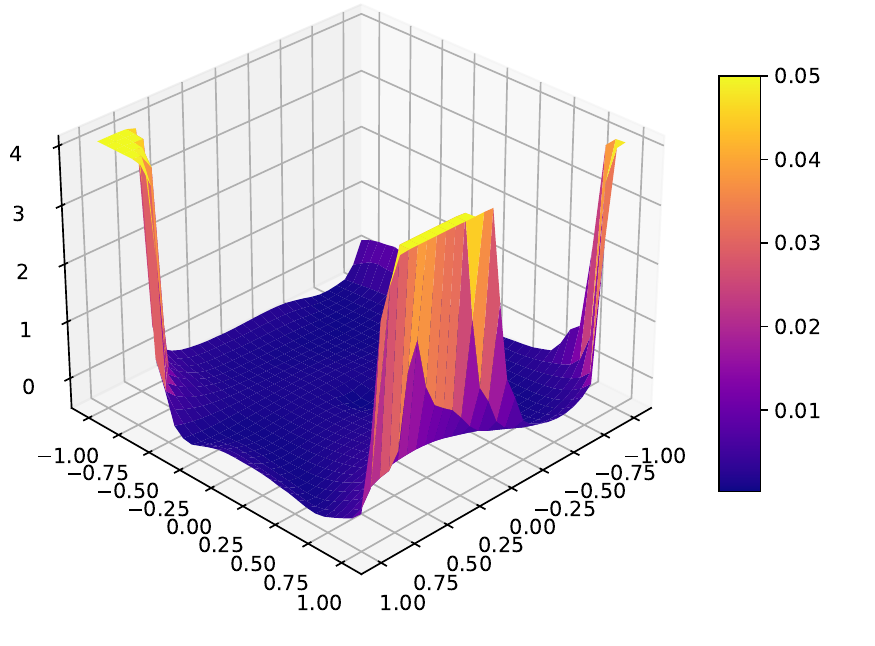}
    \subcaption{DyGFormer}
  \end{subfigure}%
  \begin{subfigure}[b]{0.248\textwidth}
    \includegraphics[width=\linewidth]{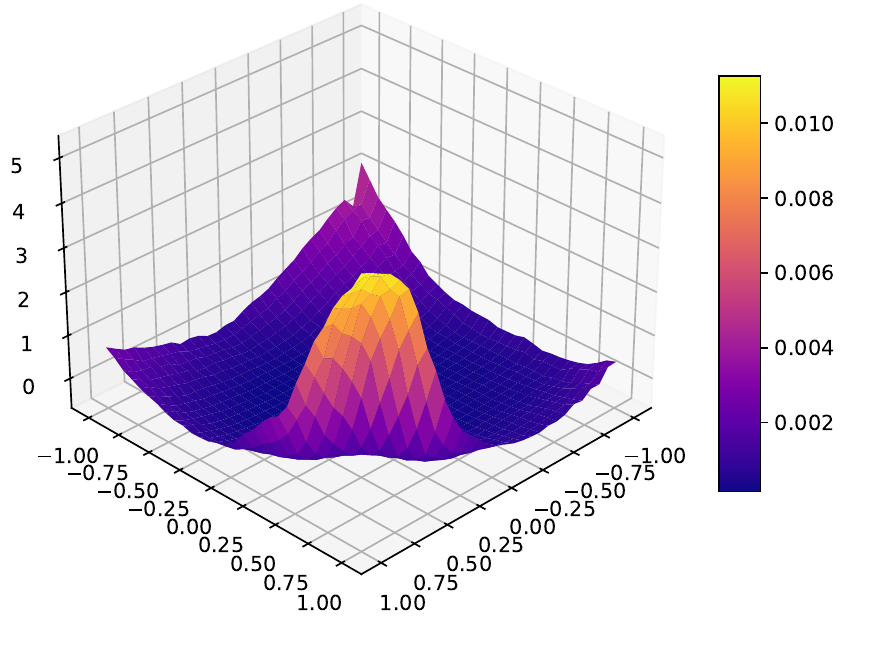}
    \subcaption{GRN}
  \end{subfigure}
\caption{2D (Filled Contour) and 3D (Surface) loss landscape on Wikipedia dataset. We show the top-ranked performers here, and additional loss landscapes for all baselines are in Appendix \ref{appendix:losslandscape}.} 
\label{fig:losslandscape}
\end{figure*}

\begin{figure*}[htbp]
  \centering
  \begin{minipage}[t]{0.58\linewidth}
    \centering
    \includegraphics[width=\linewidth]{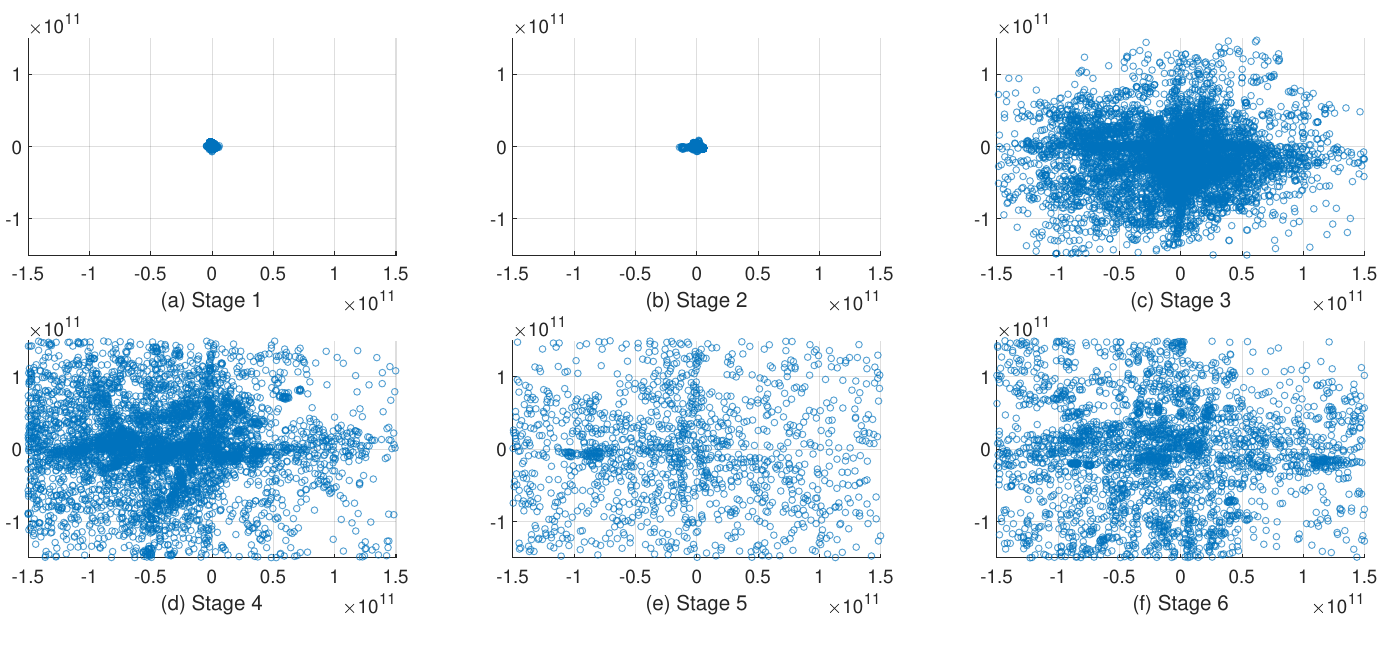}
    \captionof{figure}{T-SNE visualization of recurrent states with $\lambda = 1 - 2^{-6}$. Each stage represents the same 5\% interval of the input data within a training epoch.}
    \label{fig:state:lambda1}
  \end{minipage}%
  \hfill
  \begin{minipage}[t]{0.38\linewidth}
    \centering
    \includegraphics[width=\linewidth]{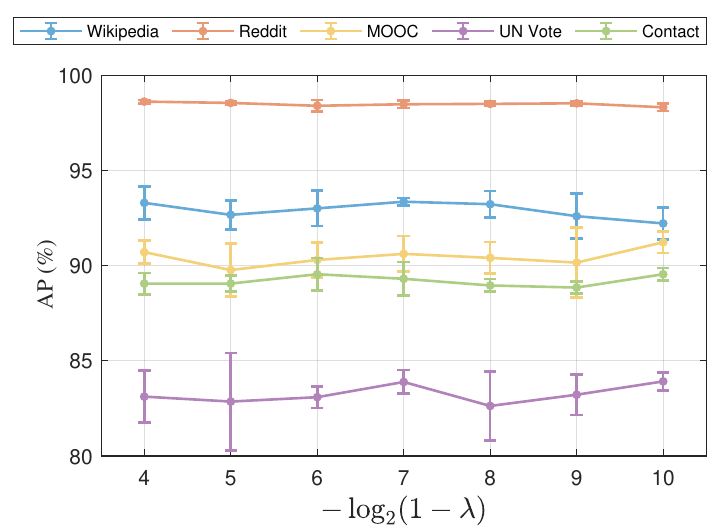}
    \captionof{figure}{Impact of decay strength with 95\% CI on AP across five datasets under transductive setting.}
    \label{fig:decay_trans}
  \end{minipage}
\end{figure*}

\textbf{Effectiveness.} \textit{The \our{} achieves strong and comprehensive performance} in both transductive and inductive settings (see Tables~\ref{tab:transAP},\ \ref{tab:transAUC}, \ref{tab:inducAP}, and \ref{tab:inducAUC}), as well as on link prediction and node classification tasks (see Table~\ref{tab:nodeAUC}). The results highlight the \our{}'s robustness to a wide range of temporal graph structures. On CTDGs such as Wikipedia, Reddit, and MOOC, the \our{} consistently ranks among the top performers, demonstrating its ability to model irregular and fine-grained temporal patterns. It also excels on long-term and sparse graphs like UN Vote and US Legis., where balancing long-term and short-term dependencies is critical. In contrast, for dense graphs with frequent short-interval interactions such as Contact and Flights, the \our{} maintains competitive performance, showcasing its scalability and efficiency. Moreover, its strong results on imbalanced node classification tasks with limited features (see Appendix~\ref{appendix:node}) further attest to its architectural generality. These empirical gains stem from three key factors: (\romannumeral1) the recurrent state $\mathbf{S}$ captures long-term temporal dependencies, while the decay factor $\lambda$ conjugates with it to allow information decay, enhancing temporal sensitivity (see Section~\ref{section-5:whygraphretention}); (\romannumeral2) the graph retention mechanism yields a smooth optimization landscape conducive to generalization (see Figure~\ref{fig:losslandscape}); and (\romannumeral3) the unified design supports versatile learning across both DTDGs and CTDGs without requiring model reconfiguration (see Section~\ref{ablation}). However, we find some performance differences between the \our{} and feature-enhanced baselines on relatively dense graphs (e.g. LastFM and Contact). Unlike methods that improve performance through input-level augmentation (e.g., neighbor co-occurrence encoding and patching in DyGFormer), GRN is designed as a unified, end-to-end architecture that learns directly from raw input features. To better understand the performance gap, we conduct an exploratory comparison by applying the identical feature enhancement strategies to GRN. As shown in Table~\ref{tab:grnplus}, this variant (denoted as GRN+) narrows the gap, achieving up to +23.31\% improvement on transductive LastFM and consistent gains across both evaluation settings. 
\begin{table}[!htb]
\centering
\caption{Comparison of predictive performance with feature enhancement (GRN+).}
\label{tab:grnplus}
\resizebox{1\linewidth}{!}{
\begin{tabular}{l|l|cccccc}
\toprule
Setting & Dataset & CAWN & GraphMixer & DyGFormer & GRN & GRN+ & Improvement \\ \midrule
Transductive & LastFM & 83.09$\pm$0.05 & 69.64$\pm$0.03 & 90.05$\pm$0.03 & 72.93$\pm$0.62 & 89.93$\pm$0.05 & \textbf{+23.31\%} \\
             & Contact & 87.26$\pm$0.20 & 89.28$\pm$0.31 & 97.89$\pm$0.05 & 89.34$\pm$0.75 & 97.93$\pm$0.08 & \textbf{+09.61\%} \\ \midrule
Inductive    & LastFM & 87.55$\pm$0.09 & 81.63$\pm$0.20 & 93.11$\pm$0.03 & 83.44$\pm$0.95 & 91.04$\pm$0.15 & \textbf{+09.11\%} \\
             & Contact & 86.36$\pm$0.31 & 87.46$\pm$0.31 & 97.71$\pm$0.04 & 91.18$\pm$1.12 & 98.09$\pm$0.03 & \textbf{+07.58\%} \\ \bottomrule
\end{tabular}}
\end{table}
\textbf{Scalability.}
The proposed \our{} architecture demonstrates strong scalability, as evidenced by its consistent superiority across all experimental settings in both \textit{efficiency} and \textit{effectiveness}. To better understand its generalizability, we visualize the 2D (filled contour) and 3D (surface) loss landscapes of the top-performing models on the Wikipedia dataset in Figure~\ref{fig:losslandscape}. Compared to competitive baselines, the \our{} displays a distinctly flatter and broader minimum in both 2D and 3D spaces. Additional results on Reddit and MOOC datasets are provided in Appendix~\ref{appendix:losslandscape}, which further substantiate the generalization advantage of \ours{} under varied graph densities and event granularities.

\subsection{Why is Graph Retention Well-Suited for Dynamic Graph Modeling?}
\label{section-5:whygraphretention}
While comparative experiments on the efficiency, effectiveness, and scalability of \ours{} demonstrate its excellent ability to model dynamic graphs, we further discuss \textit{why it is well-suited for dynamic graph modeling} in terms of its core graph retention operator.

\textbf{Recurrent states contribute to dynamic graph representation learning.}
To better understand how recurrent states contribute to dynamic graph representation learning, we perform detailed t-SNE visualizations of recurrent states over six chronological stages under different scale decay factors in Figures~\ref{fig:state:lambda1} and \ref{fig:state:lambda2}. Across all settings, we observe a consistent dynamic unfolding of the latent states. At early stages (Stage 1–2), recurrent states are tightly clustered at the presented order of magnitude, indicating relatively minimal temporal differentiation. As time progresses (Stage 3–6), the state representations gradually disperse and form more structured and heterogeneous clusters. This expansion clearly reflects the accumulation of temporal signals, highlighting the ability of the recurrent state to encode long-term interactions. The visualizations provide strong empirical evidence that the recurrent state encourages temporally smooth yet discriminative latent dynamics, supporting the \our{}'s ability to generalize.

\textbf{Multi-scale decay balances long-term and short-term dependencies.}
The multi-scale decay mechanism in \ours{} enables the model to flexibly adapt to datasets with highly diverse temporal dynamics. As shown in Figures~\ref{fig:decay_trans} and~\ref{fig:decay_induc}, AP scores vary  with different decay levels across both transductive and inductive link prediction tasks on most datasets. For datasets characterized by dense and rapidly evolving interactions (Contact and MOOC), stronger decay generally leads to better performance by emphasizing recent events and reducing the influence of outdated information. In contrast, for datasets with sparse or long-term temporal dependencies (UN Vote), weaker decay is often more effective, allowing the model to retain and leverage long-term temporal signals. These results further underscore the benefit of multi-head temporal modeling, where assigning distinct decay factors to each head allows \ours{} to concurrently capture both short-term and long-term patterns.

\subsection{Ablation Study}
\label{ablation}
To evaluate the contribution of individual components within the architecture, we ablate various design choices of \our{}. To isolate the effect of each design component, we construct the following \our{} variants:
\begin{itemize}
    \item w/o $\mathbf{S}$ recurrent state: removes the recurrent temporal memory module, forcing the model to rely solely on instantaneous observations.
    \item w/o $\lambda$ decay: disables the temporal decay operator, treating all past interactions equally regardless of recency (i.e. $\lambda=1$).
    \item w/o multi-scale decay: replaces adaptive decay with a fixed, single-scale decay, reducing flexibility across datasets.
    \item w/o temporal encoding: removes time encoding, making the model invariant to the order and spacing of temporal events.
\end{itemize}

\begin{figure}[!htb]
\label{fig:variants}
\centering
\begin{subfigure}[]{0.9\columnwidth}
\centering
\includegraphics[width=\columnwidth]{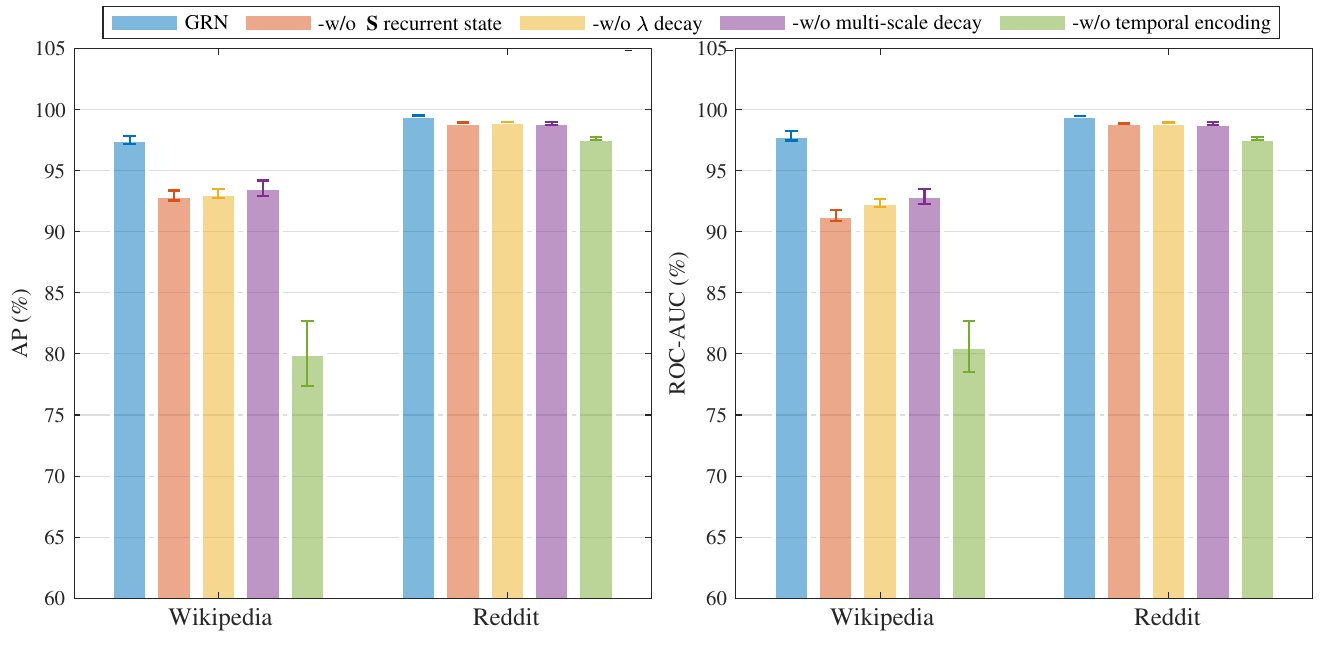}
\caption{Transductive performance.}
\end{subfigure}
\hfill
\begin{subfigure}[]{0.9\columnwidth}
\centering
\includegraphics[width=\columnwidth]{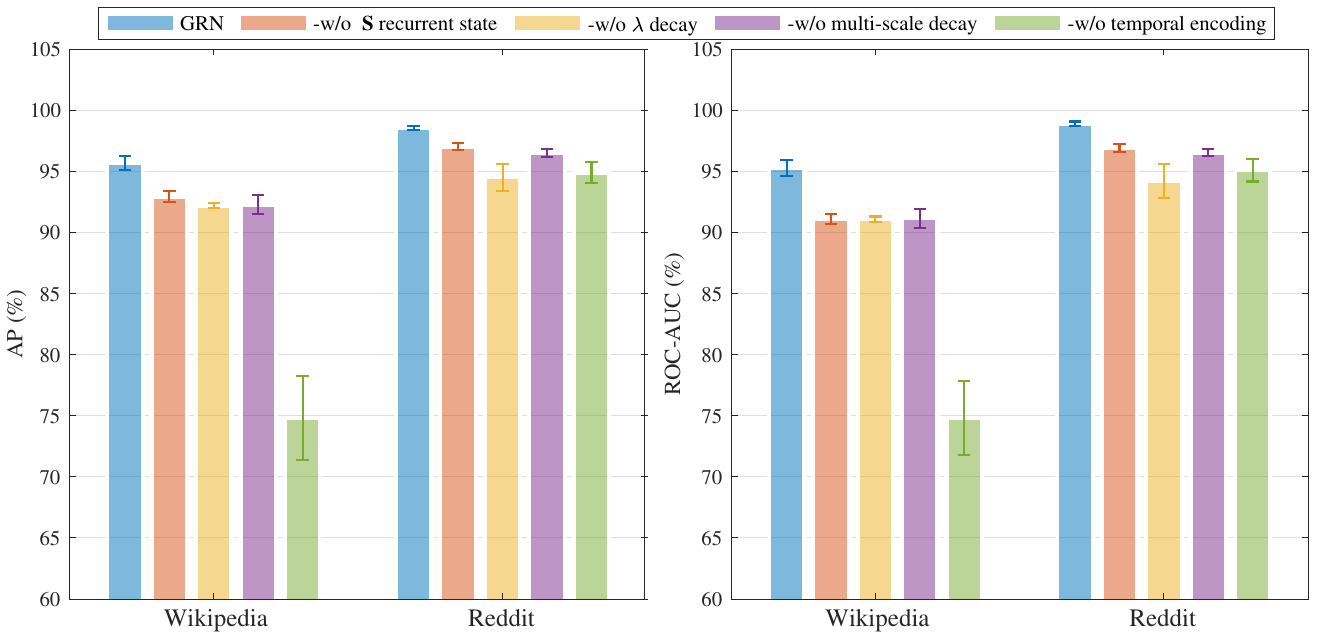}
\caption{Inductive performance.}
\end{subfigure}
\caption{Results of ablation experiments on GRN components.}
\label{fig:ablation}
\end{figure}

The results in Figure~\ref{fig:ablation} offer a systematic breakdown of each architectural component’s contribution to the \our{} under both transductive and inductive link prediction settings on the Wikipedia and Reddit datasets. Overall, the ablation study confirms the complementarity and necessity of all architectural components. The recurrent state $\mathbf{S}$ enables long-term memory, the decay module provides time-sensitive signal weighting, and temporal encoding allows the model to resolve precise temporal order. Their synergy is essential for achieving strong generalization in both seen (transductive) and unseen (inductive) dynamic graph settings. Detailed analysis is provided in Appendix~\ref{appendix:ablation}

\section{Conclusion}
\label{section-6}
In this work, we propose the \our{} as a unified and scalable architecture that incorporates a novel graph retention mechanism to effectively address key challenges in dynamic graph learning. Extensive experiments across a wide range of dynamic graph tasks and settings confirm its high efficiency, comprehensive effectiveness, and strong scalability, showcasing its robustness and versatility across various dynamic graph tasks. We believe this work opens up promising directions for further exploration in dynamic graph representation learning. Concerns and limitations are discussed in Appendix~\ref{appendix:lims}, which we aim to address in future work.

\section*{ACKNOWLEDGMENTS}
This project was supported by the NZ EQUALS program (4260/76892); the Post-funded Key Project of National Social Science Fund of China (25FGLA001); the Ministry of Education of Humanities and Social Science Project of China (25YJA630043); the Natural Science Foundation of Hubei Province of China (JCZRYB202401041); the National Natural Science Foundation of China (71974069); the Program of China Scholarship Council (202408250021 and 202506770023).

\bibliographystyle{ACM-Reference-Format}
\balance
\bibliography{references}

\appendix
\label{section-appendix}

\section{Additional Descriptions}
\label{additionaldescriptions}

\subsection{Time and Space Complexity Analysis}
\label{appendix:complexity}

\begin{algorithm}[!ht]
\caption{Overall process of \ours{}}
\label{algorithm}
\SetAlgoNlRelativeSize{1}
\SetAlgoNlRelativeSize{0}
\SetAlgoNlRelativeSize{-1}
\SetAlgoNlRelativeSize{-2}
\SetAlgoNlRelativeSize{-3}
\KwIn{Dynamic graph $\mathcal{G} = (\mathcal{V}, \mathcal{E})$, node features $\mathbf{X}$, edge attributes $\mathbf{E}$, time intervals $\Delta \mathbf{T}$}
\KwOut{Node representations $\mathbf{O}$}

\SetKwInOut{Input}{Input}
\SetKwInOut{Output}{Output}

\SetKwFunction{FMessage}{Message}
\SetKwFunction{FGraphRet}{GraphRetention}
\SetKwFunction{FMultiHead}{MultiHeadRetention}
\SetKwFunction{FFFN}{FFN}
\SetKwFunction{FConcat}{Concatenate}
\SetKwFunction{FNormalize}{GroupNorm}

\textbf{Initialize:} All trainable parameters $\mathbf{W,b}$ for each head $h$; Recurrent state $\mathbf{S}_0$\;

\For{$t = 1$ \textbf{to} $T$}{
    \For{each node $v_t^i$ (destination node)}{
        
        $\mathbf{X}^{j}_t \leftarrow \FMessage(\mathbf{X}^{j}_t, \mathbf{E}^{j}_t, \Delta \mathbf{T}_t)$\;

        \For{each head $h = 1, \dots, H$}{
            \texttt{\# Compute Q, K, V with head-specific parameters}\;
            $
                \mathbf{Q}_{t,h} = \mathbf{x}_t^i \mathbf{W}_{q,h} + \mathbf{1}^\top \mathbf{b}_{q,h}
            $\;
            $
                \mathbf{K}_{t,h} = \mathbf{X}_t^j \mathbf{W}_{k,h} + \mathbf{1}^\top \mathbf{b}_{k,h}
            $\;
            $
                \mathbf{V}_{t,h} = \mathbf{X}_t^j \mathbf{W}_{v,h} + \mathbf{1}^\top \mathbf{b}_{v,h}
            $

            \texttt{\# Compute graph retention output for each head}\;
            $
                \mathbf{o}_{t,h}^i = \FGraphRet(\mathbf{Q}_{t,h}, \mathbf{K}_{t,h}, \mathbf{V}_{t,h}, \mathbf{S}_{t-1})
            $
        }
    }
    \texttt{\# Concatenate all head outputs}\;
    $\mathbf{H} = \FConcat(\mathbf{o}_{t,1}^i, \dots, \mathbf{o}_{t,H}^i)$ \;
    \texttt{\# Normalize the concatenated output}\;
    $\mathbf{H} \leftarrow \FNormalize(\mathbf{H})$\;

    \texttt{\# Feed forward}\;
    $
        \mathbf{O} = \FFFN(\mathbf{H})
    $

    \texttt{\# Update destination nodes' features}\;
    $
        \mathbf{X}_{t+1} \leftarrow \mathbf{O}
    $
}
\end{algorithm}

The computational efficiency of GRN arises from its flexible retention paradigms and parallelizable operations. Below, we analyze the time and space complexity of each core component.

\textit{Message and Temporal Encoding.}
Each node aggregates temporal messages from its historical neighbors. The message function requires $\mathcal{O}(d)$ operations per edge for linear transformations and temporal encoding. Given $|E_t|$ edges at time $t$, this step incurs a total cost of $\mathcal{O}(|E_t| \cdot d)$ per time step.

\textit{Graph Retention Operator.}
For a destination node with $|\mathcal{N}(i)|$ historical neighbors:
\begin{itemize}
  \item In the parallel paradigm, computing $\mathbf{QK}^\top$ where $\mathbf{Q}, \mathbf{K} \in \mathbb{R}^{|\mathcal{N}(i)| \times d}$ costs $\mathcal{O}(|\mathcal{N}(i)|^2 \cdot d)$, followed by multiplication with $\mathbf{V}$. Across all nodes, the total complexity is $\mathcal{O}(|V_t| \cdot \bar{N}^2 \cdot d)$.
  \item In the recurrent paradigm, the state $\mathbf{S}_t \in \mathbb{R}^{d \times d}$ is updated via $\mathbf{S}_t = \lambda \mathbf{S}_{t-1} + \mathbf{K}_t^\top \mathbf{V}_t$. Each update costs $\mathcal{O}(d^2)$, yielding $\mathcal{O}(|V_t| \cdot \bar{N} \cdot d^2)$ in total.
  \item The chunkwise paradigm processes neighbors in chunks of size $B$. Intra-chunk attention costs $\mathcal{O}(B^2 \cdot d)$, cross-chunk state costs $\mathcal{O}(B \cdot d^2)$. Total: $\mathcal{O}(|V_t| \cdot \bar{N} \cdot (B \cdot d + d^2))$, with peak memory reduced to $\mathcal{O}(B^2 + d^2)$.
\end{itemize}

\textit{Multi-Scale Graph Retention.}
Each of $H$ heads executes graph retention independently: $\mathcal{O}(H \cdot \bar{N}^2 \cdot d)$ (parallel) or $\mathcal{O}(H \cdot \bar{N} \cdot d^2)$ (recurrent). Group normalization adds $\mathcal{O}(H \cdot d)$.

\textit{Feedforward Network (FFN).}
Two linear layers with $d_{model} = H \cdot d$. Complexity: $\mathcal{O}(d_{model}^2)$ per node.

\textit{Overall Complexity.}
With $L$ layers, total time complexity per step:
$\mathcal{O}(L \cdot |V_t| \cdot (H \cdot \bar{N}^2 \cdot d + d_{model}^2))$ (parallel); $\mathcal{O}(L \cdot |V_t| \cdot (H \cdot \bar{N} \cdot d^2 + d_{model}^2))$ (recurrent).
Space complexity: $\mathcal{O}(|V_t| \cdot L \cdot H \cdot d^2 + |V_t| \cdot d_{model} + |E_t| \cdot d)$.
The recurrent paradigm is advantageous when $\bar{N} > d$, avoiding $\mathcal{O}(\bar{N}^2)$ attention maps for long-range graphs.

\begin{table*}[!htb]
\centering
\caption{Statistics of datasets. Type: D = DTDG, C = CTDG; N.\&F. Feat. = Node \& Link Feature dimension. Graphs are classified as DTDGs if the number of links significantly exceeded the number of unique timesteps, as interactions within each batch primarily occurred within one single temporal snapshot.}
\label{tab:datasets}
\resizebox{2\columnwidth}{!}{
\begin{tabular}{l|ccccccccc}
\hline
\toprule
Datasets  & Type & Domains   & Bipartite & \#Nodes & \#Links & \#N.\&L. Feat. & \#Unique Steps & Duration   & Time Granularity \\ \midrule
Wikipedia  & C  & Social   & True   & 9227  & 157474 & - \& 172    & 152757    & 1 month    & Unix timestamps \\
Reddit   & C  & Social   & True   & 10984  & 672447 & - \& 172    & 669065    & 1 month    & Unix timestamps \\
MOOC    & C  & Interaction & True   & 7144  & 411749 & - \& 4     & 345600    & 17 months   & Unix timestamps \\
LastFM   & C  & Interaction & True   & 1980  & 1293103 & - \& -     & 1283614   & 1 month    & Unix timestamps \\
Myket    & C  & Interaction & True   & 17988  & 694121 & - \& -     & 693774    & 197 days   & Unix timestamps \\
Enron    & C  & Social   & False   & 184   & 125235 & - \& -     & 22632    & 3 years    & Unix timestamps \\
UCI     & C  & Social   & False   & 1899  & 59835  & - \& -     & 58911    & 196 days   & Unix timestamps \\
Flights   & D  & Transport  & False   & 13169  & 1927145 & - \& 1     & 122     & 4 months   & days       \\
Can. Parl. & D  & Politics  & False   & 734   & 74478  & - \& 1     & 14      & 14 years   & years      \\
US Legis.  & D  & Politics  & False   & 225   & 60396  & - \& 1     & 12      & 12 congresses & congresses    \\
UN Trade  & D  & Economics  & False   & 255   & 507497 & - \& 1     & 32      & 32 years   & years      \\
UN Vote   & D  & Politics  & False   & 201   & 1035742 & - \& 1     & 72      & 72 years   & years      \\
Contact   & C  & Proximity  & False   & 692   & 2426279 & - \& 1     & 8064     & 1 month    & 5 minutes    \\
\bottomrule
\end{tabular}}
\end{table*}

\subsection{Additional Descriptions of Datasets}
\label{appendix:dataset}
The statistics for the datasets used are summarized in Table~\ref{tab:datasets}.

\subsection{Descriptions of Baselines}
\label{appendix:baseline}

The details of the nine baselines are described as follows:

\begin{itemize}
\item \textbf{JODIE}: Designed for temporal bipartite networks such as user-item interactions, JODIE utilizes two coupled recurrent neural networks to dynamically update the states of users and items upon interactions. A projection mechanism predicts future user/item embeddings by learning their temporal trajectories.

\item \textbf{TGAT}: Utilizes graph attention mechanisms combined with temporal encoding to model node dynamics over time. This approach ensures effective representation learning by combining temporal and structural graph information.

\item \textbf{TGN}: Incorporates a memory module for each node to maintain evolving states, which are updated through a message-passing paradigm consisting of a message function, aggregator, and updater. An embedding module generates temporal node representations.

\item \textbf{CAWN}: Employs causal Anonymous Walks to capture causal patterns in dynamic graphs. Each walk is encoded using recurrent neural networks, and the results are aggregated to form node representations. This method effectively leverages temporal and causal graph structures.

\item \textbf{EdgeBank}: A memory-based model for dynamic link prediction, it retains observed edges in memory and predicts interactions based on past edge information. Variants such as $\text{EdgeBank}_{\infty}$ and $\text{EdgeBank}_{\text{tw}}$ adapt memory retention to different temporal contexts, such as fixed-size windows or edge appearance thresholds.

\item \textbf{TCL}: Constructs temporal subgraphs using a breadth-first search. It applies a graph transformer that integrates graph topology and temporal attributes while modeling interdependencies through cross-attention mechanisms.

\item \textbf{GraphMixer}: Introduces fixed time encoding and uses an MLP-Mixer architecture for link and node features to create effective representations. This model emphasizes simplicity while ensuring robustness across temporal graph tasks.

\item \textbf{DyGFormer}: Incorporates attention mechanisms to handle dynamic graph structures and temporal sequences. Its architecture includes multiple transformer layers and employs co-occurrence and aligned encoding to enhance node representations. The model supports varying sequence lengths and dynamic contexts, offering high adaptability.

\end{itemize}

\subsection{Detailed Experimental Configurations and Hyperparameters}
\label{appendix:configurations}
We adopt a 70\%-15\%-15\% chronological train-validation-test split, with 10\% of nodes designated as unobserved to evaluate inductive performance. The model comprises 2 GRN blocks with 2 attention heads, using an embedding dimension of 64 for both node and temporal representations. Training is performed using the Adam optimizer with binary cross-entropy loss, a learning rate of $10^{-4}$, and a batch size of 200. The noise schedule parameter $\lambda$ is computed as $1 - \exp(\mathrm{linspace}(\ln 2^{-6}, \ln 2^{-10}, 2))$. Layer-wise dropout rates are set to $[0.1, 0.3, 0.2, 0.1, 0.1, 0.2, 0.1, 0.1, 0.1, 0.1, 0.1, 0.1, 0.1]$, with optimal hyperparameters selected via grid search. Each model is trained for 50 epochs with early stopping (patience $= 10$) and evaluated over five runs with different random seeds initialized from system timestamps. Detailed baseline configurations are available in our code repository. All experiments are conducted on Ubuntu with an NVIDIA GeForce RTX 4090 GPU (24GB) and an Intel Core i9-13900KF CPU.

\subsection{Clarification on Inconsistencies in Experimental Implementation and Settings}
\label{appendix:implementationclarification}

\paragraph{Inconsistent Evaluation Protocol for Inductive Setting}
We note that the DyGLib follows the same inductive evaluation protocol as TGAT~\footnote{\url{https://github.com/StatsDLMathsRecomSys/Inductive-representation-learning-on-temporal-graphs}} and TGN~\footnote{\url{https://github.com/twitter-research/tgn}}, where unseen nodes are included in the validation and test sets under the transductive setting. In contrast, our experiments follow the protocol proposed by CAWN~\footnote{\url{https://github.com/snap-stanford/CAW}}, ensuring that unseen nodes are strictly excluded from the transductive setup. Such discrepancies, particularly in the inclusion of unseen nodes during transductive evaluation and the use of early stopping, introduce inconsistencies in the comparative results of both transductive and inductive evaluations.

\section{Additional Experimental Results}
\label{appendix:Additional Experimental Results}

\subsection{Edge-Level Prediction Performance}
\label{appendix:link}
We report the results of the comparison experiments with baselines for the link prediction task, presenting ROC-AUC scores under the transductive setting in Table~\ref{tab:transAUC}, and both AP and ROC-AUC scores under the inductive setting in Tables~\ref{tab:inducAP} and ~\ref{tab:inducAUC}.

\begin{table}[!htb]
\centering
\caption{AUC-ROC for transductive link prediction on dynamic graphs with random negative sampling strategy.}
\label{tab:transAUC}
\resizebox{1\columnwidth}{!}{
\begin{tabular}{l|cccccccccc}
\hline
\toprule
Datasets & JODIE & DyRep & TGAT & TGN & CAWN & EdgeBank & TCL & GraphMixer & DyGFormer & GRN \\
\midrule
Wikipedia & 91.48±3.34 & 90.60±0.59 & 93.36±0.35 & 97.25±0.07 & \underline{98.05±0.03} & 90.78±0.00 & 94.45±0.10 & 95.68±0.05 & \textbf{98.68±0.03} & 97.86±0.40 \\ 
Reddit & 97.56±0.16 & 97.42±0.12 & 96.58±0.04 & 96.93±0.16 & 98.63±0.03 & 95.20±0.00 & 93.21±0.17 & 95.26±0.04 & \underline{98.86±0.06} & \textbf{99.47±0.03} \\ 
MOOC & \underline{81.22±4.43} & 74.20±1.09 & 78.97±1.69 & 76.45±1.29 & 70.29±0.81 & 60.86±0.00 & 76.02±1.85 & 75.23±0.27 & 72.39±1.03 & \textbf{92.89±1.06} \\ 
LastFM & 69.91±0.66 & 69.79±1.41 & 63.03±0.33 & 53.33±1.75 & 81.44±0.10 & \underline{83.77±0.00} & 56.36±0.19 & 68.28±0.02 & \textbf{90.10±0.04} & 72.47±0.62 \\ 
Myket & \textbf{86.37±0.12} & \underline{86.35±0.13} & 66.04±5.60 & 82.69±2.08 & 83.72±0.16 & 57.35±0.00 & 68.81±2.18 & 85.00±0.04 & 78.99±0.39 & 82.36±1.94 \\ 
Enron & 83.27±0.65 & 75.67±1.41 & 65.10±0.53 & 75.35±2.50 & 85.44±0.90 & \textbf{87.05±0.00} & 67.52±1.53 & 84.12±0.14 & \underline{86.31±2.44} & 84.50±2.87 \\ 
UCI & 83.66±1.13 & 56.73±4.00 & 76.08±0.68 & 85.87±0.94 & \underline{91.78±0.23} & 77.30±0.00 & 84.37±0.78 & 89.77±0.75 & \textbf{93.94±0.11} & 80.74±1.54 \\ 
Flights & 93.59±0.42 & 91.49±0.83 & 91.63±0.02 & 93.28±1.55 & \underline{96.57±0.29} & 90.23±0.00 & 90.45±0.03 & 90.44±0.00 & \textbf{98.74±0.04} & 96.04±1.33 \\ 
Can. Parl. & 82.76±0.52 & 75.24±1.14 & 72.31±1.27 & 75.47±1.57 & 72.62±3.97 & 64.14±0.00 & 69.59±3.44 & 81.66±1.11 & \textbf{94.93±1.28} & \underline{90.47±0.13} \\ 
US Legis. & \underline{85.62±0.14} & 75.42±2.36 & 64.66±3.96 & 77.92±0.27 & 70.36±2.76 & 62.57±0.00 & 60.75±0.99 & 77.77±0.56 & 78.94±0.27 & \textbf{86.38±0.32} \\ 
UN Trade & \underline{71.04±0.48} & 69.03±1.09 & 62.87±0.44 & 65.47±0.26 & 65.32±0.11 & 66.75±0.00 & 63.10±0.17 & 54.17±6.33 & 60.00±2.14 & \textbf{72.33±0.52} \\ 
UN Vote & \underline{68.72±1.60} & 65.45±0.88 & 52.78±0.62 & 61.12±1.43 & 52.33±0.06 & 62.97±0.00 & 50.88±0.20 & 52.37±0.34 & 54.50±0.12 & \textbf{86.01±0.97} \\ 
Contact & 95.00±0.15 & 88.53±1.56 & \underline{95.11±0.07} & 94.18±1.41 & 85.31±0.19 & 94.34±0.00 & 92.27±0.22 & 92.25±0.24 & \textbf{98.15±0.05} & 91.02±0.59 \\ 
\midrule
Avg. Rank & 3.69 & 6.23 & 7.08 & 5.38 & 5.15 & 6.92 & 7.85 & 6.15 & \underline{3.31} & \textbf{3.23} \\
\bottomrule
\end{tabular}}
\end{table}

\begin{table}[!htb]
\centering
\caption{AP for inductive link prediction on dynamic graphs with random negative sampling strategy.}
\label{tab:inducAP}
\resizebox{1\columnwidth}{!}{
\begin{tabular}{l|ccccccccc}
\hline
\toprule
Datasets  & JODIE   & DyRep   & TGAT    & TGN    & CAWN    & TCL    & GraphMixer & DyGFormer & GRN         \\ 
\midrule
Wikipedia & 86.38±4.17 & 84.29±0.88 & 93.91±0.24 & 96.95±0.12 & \underline{97.97±0.01} & 95.17±0.11 & 95.41±0.09 & \textbf{98.35±0.07} & 95.69±0.57 \\ 
Reddit & 91.95±0.45 & 88.36±0.42 & 93.70±0.15 & 90.50±1.60 & 98.16±0.03 & 88.66±0.12 & 92.46±0.04 & \underline{98.46±0.04} & \textbf{98.57±0.16} \\ 
MOOC & 70.50±2.51 & 64.37±2.06 & \underline{74.70±2.23} & 74.16±3.33 & 64.71±0.97 & 71.14±1.51 & 68.35±0.65 & 64.39±0.94 & \textbf{90.59±0.77} \\ 
LastFM & 82.65±1.02 & 82.36±1.93 & 75.17±0.20 & 65.89±6.42 & \underline{87.55±0.09} & 71.85±0.20 & 81.63±0.20 & \textbf{93.11±0.03} & 83.44±0.95 \\ 
Myket & \textbf{80.58±0.38} & 76.84±0.94 & 60.99±4.85 & 61.30±5.02 & 77.76±0.24 & 63.27±2.72 & \underline{80.50±0.03} & 72.90±0.57 & 73.23±4.91 \\ 
Enron & 75.38±0.87 & 62.02±4.84 & 66.30±0.67 & 68.25±1.70 & \underline{82.58±0.57} & 67.73±2.17 & 74.44±0.48 & \textbf{83.96±2.06} & 81.39±2.94 \\ 
UCI & 73.94±1.78 & 49.55±1.87 & 79.70±0.60 & 81.62±0.68 & 89.80±0.20 & 82.00±1.05 & \underline{90.16±0.47} & \textbf{93.78±0.07} & 75.27±1.35 \\ 
Flights & 86.61±1.82 & 82.88±1.20 & 83.64±0.12 & 87.47±2.45 & \underline{94.56±0.35} & 79.96±0.47 & 79.79±0.18 & \textbf{97.43±0.05} & 87.55±3.76 \\ 
Can. Parl. & 51.84±1.17 & 50.58±1.18 & 53.35±0.41 & 53.42±0.37 & 55.13±1.29 & 53.48±0.88 & 57.39±0.35 & \textbf{81.86±0.69} & \underline{77.58±1.90} \\ 
US Legis. & 52.31±0.32 & 55.75±1.06 & 55.77±4.45 & \underline{59.69±0.15} & 59.52±0.73 & 59.22±0.40 & 53.89±1.47 & 55.49±0.54 & \textbf{77.52±4.37} \\ 
UN Trade & 58.55±0.56 & 56.44±0.50 & 59.72±0.93 & 52.70±0.71 & \underline{63.29±0.16} & 61.12±0.66 & 54.10±4.47 & 55.72±1.74 & \textbf{64.26±2.15} \\ 
UN Vote & 50.50±1.30 & 51.08±1.55 & 50.79±0.43 & 49.91±1.54 & 49.09±0.85 & 52.72±1.08 & \underline{54.43±0.43} & 53.64±0.21 & \textbf{78.07±1.38} \\ 
Contact & \underline{93.14±0.63} & 76.31±9.60 & 92.68±0.22 & 87.04±3.75 & 86.36±0.31 & 87.38±0.56 & 87.46±0.31 & \textbf{97.71±0.04} & 91.18±1.12 \\ 
\midrule
Avg. Rank & 5.54 & 7.38 & 5.69 & 5.92 & 3.85 & 5.69 & 5.08 & \underline{3.08} & \textbf{2.77} \\
\bottomrule
\end{tabular}
}
\end{table}

\begin{table}[!htb]
\centering
\caption{AUC-ROC for inductive link prediction on dynamic graphs with random negative sampling strategy.}
\label{tab:inducAUC}
\resizebox{1\columnwidth}{!}{
\begin{tabular}{l|ccccccccc}
\hline
\toprule
Datasets  & JODIE   & DyRep   & TGAT    & TGN    & CAWN    & TCL    & GraphMixer & DyGFormer & GRN         \\ 
\midrule
Wikipedia & 85.97±3.36 & 82.93±0.69 & 93.44±0.28 & 96.69±0.14 & \underline{97.59±0.02} & 94.62±0.07 & 95.24±0.08 & \textbf{98.19±0.06} & 95.27±0.65 \\ 
Reddit & 91.43±0.43 & 87.81±0.56 & 93.85±0.15 & 91.40±1.19 & 97.78±0.04 & 89.11±0.12 & 92.39±0.03 & \underline{98.21±0.07} & \textbf{98.90±0.17} \\ 
MOOC & 75.66±2.78 & 67.86±1.77 & 76.05±2.85 & \underline{76.07±3.82} & 65.62±0.92 & 71.98±2.29 & 70.00±0.36 & 65.56±1.02 & \textbf{91.82±0.69} \\ 
LastFM & 82.12±1.09 & 81.64±1.98 & 74.67±0.18 & 64.02±5.23 & \underline{85.07±0.14} & 69.83±0.18 & 80.35±0.05 & \textbf{92.87±0.05} & 82.11±1.04 \\ 
Myket & \textbf{77.71±0.45} & 74.48±1.22 & 60.37±5.60 & 59.88±6.34 & 74.54±0.22 & 60.91±2.46 & \underline{77.67±0.06} & 69.57±0.53 & 72.60±5.53 \\ 
Enron & 76.53±1.34 & 62.37±6.40 & 64.13±0.96 & 67.50±1.46 & \underline{82.16±0.55} & 65.46±2.49 & 75.70±0.60 & \textbf{83.12±2.57} & 78.95±3.15 \\ 
UCI & 75.25±1.09 & 49.14±2.81 & 76.97±0.77 & 78.27±0.70 & 87.13±0.24 & 79.72±0.92 & \underline{88.33±0.54} & \textbf{91.54±0.11} & 72.37±1.36 \\ 
Flights & 87.16±1.40 & 82.28±1.42 & 83.67±0.08 & 87.98±3.11 & \underline{93.60±0.47} & 80.49±0.10 & 80.38±0.05 & \textbf{97.35±0.09} & 88.59±4.46 \\ 
Can. Parl. & 49.94±1.31 & 47.29±2.04 & 54.41±0.73 & 55.44±0.79 & 58.22±2.07 & 55.19±1.50 & 58.79±0.49 & \textbf{84.95±1.06} & \underline{79.15±1.14} \\ 
US Legis. & 55.41±0.33 & 58.01±1.72 & 56.27±5.33 & \underline{60.25±0.34} & 59.40±1.39 & 59.72±0.47 & 54.46±2.65 & 53.46±0.65 & \textbf{81.96±2.93} \\ 
UN Trade & 60.39±0.57 & 58.11±0.53 & 61.64±0.52 & 54.65±1.01 & \textbf{65.05±0.18} & 62.69±0.41 & 53.77±6.18 & 57.60±2.40 & \underline{63.33±2.46} \\ 
UN Vote & 47.75±2.08 & 49.91±2.26 & 50.45±0.48 & 48.77±1.91 & 47.43±0.57 & 51.54±1.15 & \underline{53.55±0.38} & 53.23±0.24 & \textbf{80.54±0.86} \\ 
Contact & 94.20±0.32 & 79.92±6.85 & \underline{94.43±0.09} & 90.95±2.42 & 84.30±0.30 & 90.93±0.23 & 90.93±0.24 & \textbf{97.96±0.04} & 91.10±0.75 \\ 
\midrule
Avg. Rank & 5.31 & 7.23 & 5.62 & 5.54 & 3.92 & 5.77 & 5.31 & \underline{3.31} & \textbf{3.00} \\
\bottomrule
\end{tabular}}
\end{table}

\subsection{Node-Level Classification Performance}
\label{appendix:node}
The datasets chosen for the node classification task are Wikipedia, Reddit and MOOC.Wikipedia has 5 classes of node states, Reddit has 15 classes of node states and MOOC has 87 classes of node states. All three datasets contain nodes without original node features. It is worth noting that the distribution of node classes in these three datasets is relatively imbalanced. We report the experimental results for the node classification task in Table~\ref{tab:nodeAUC}. The results of the comparison experiments for the node classification task are shown in Table~\ref{tab:nodeAUC}.

\begin{table}[!ht]
\centering
\caption{AUC-ROC for node classification on dynamic graphs.}
\label{tab:nodeAUC}
\resizebox{1\columnwidth}{!}{
\begin{tabular}{l|cccccccccc}
\hline
\toprule
Datasets & JODIE        & DyRep      & TGAT    & TGN    & CAWN    & TCL    & GraphMixer & DyGFormer & GRN         & \\ \midrule
Wikipedia & \underline{86.89±1.31}  & 87.72±0.71    & 68.93±3.06 & 75.82±2.60 & 79.17±1.07 & 75.85±2.18 & 72.38±2.89 & 78.99±0.12 & \textbf{88.32±1.45} & \\
Reddit  &  \textbf{64.82±1.77} & \underline{62.25±1.50} & 61.07±1.32 & 57.11±1.46 & 59.84±0.07 & 61.44±1.11 & 60.53±0.59 & 61.85±0.57 & 61.10±0.72     & \\
MOOC   & 69.04±1.27  &   64.88±2.88  &  59.92±0.72  & 70.26±1.09  &  65.24±1.06  &  67.89±0.79  &  64.19±0.59  &  \underline{73.90±0.55}  &  \textbf{76.79±2.20} &\\
\midrule
Avg. Rank & \underline{2.67} & 3.67 & 8.00 & 6.33 & 6.00 & 5.00 & 7.67 & 3.33 & \textbf{2.33}  &\\ 
\bottomrule
\end{tabular}}
\end{table}

\subsection{Additional Loss Landscape Visualizations}
\label{appendix:losslandscape}

To evaluate the generalization capability of \ours{}, we visualize and compare the loss landscapes of \ours{} and baseline models. Specifically, we give 2D filled contour plots for the Wikipedia, Reddit, and MOOC datasets in our anonymous repository. These visualizations help assess the sharpness and smoothness of the optimization landscape, which are indicative of generalization behavior.

\subsection{Additional Experimental Results of Recurrent State}
\label{appendix:recurrentstate}
\begin{figure}[htbp]
  \centering
  \begin{minipage}[t]{0.59\linewidth}
    \raggedright
    \includegraphics[width=\linewidth]{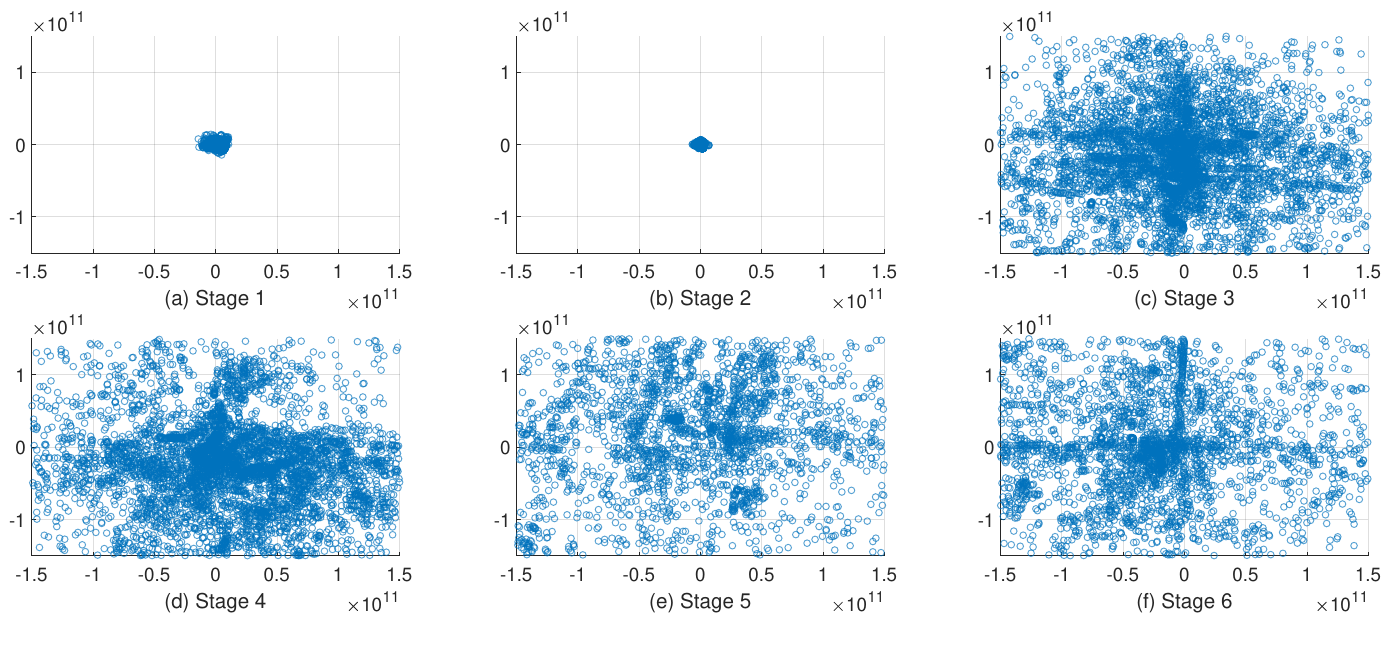}
    \captionof{figure}{T-SNE visualization of recurrent states with $\lambda = 1 - 2^{-10}$. Each stage represents the same 5\% interval of the input data within a training epoch.}
    \label{fig:state:lambda2}
  \end{minipage}%
    \hspace{0.04\linewidth} 
  \begin{minipage}[t]{0.35\linewidth}
    \raggedleft
    \includegraphics[width=\linewidth]{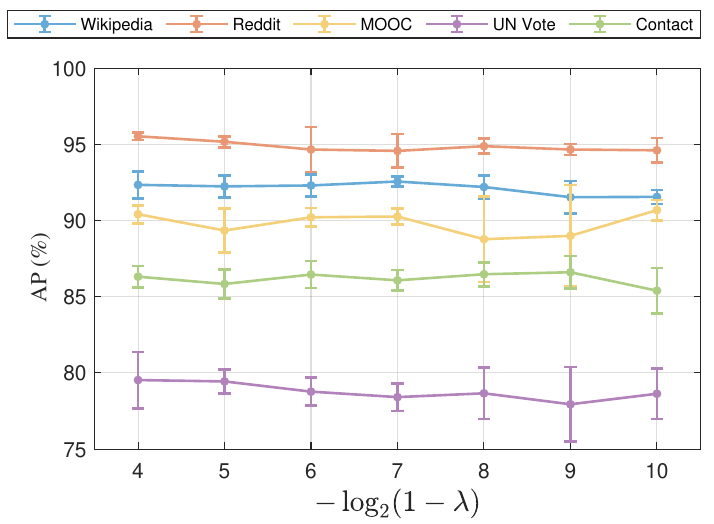}
    \captionof{figure}{Impact of decay strength with 95\% CI on AP across five datasets under inductive setting.}
    \label{fig:decay_induc}
  \end{minipage}
\end{figure}

We provide the additional results that further elucidate the contribution of the recurrent state and decay factor in Figures~\ref{fig:state:lambda2} and ~\ref{fig:decay_induc}.

\subsection{Additional Analysis of Ablation Study}
\label{appendix:ablation}

In the transductive setting, the removal of the recurrent state $\mathbf{S}$ leads to a substantial drop in both AP and ROC-AUC, especially on Wikipedia. This underscores the critical role of temporal memory in capturing evolving node-level dynamics over continuous interactions. Similarly, eliminating the decay mechanism results in noticeable performance degradation. This confirms that regulating the temporal influence of past interactions via decay is essential for learning effective temporal representations, particularly in CTDGs like Wikipedia. Interestingly, the absence of multi-scale decay only causes a marginal decline in performance, suggesting that while adaptive decay enhances robustness, a single-scale decay still retains reasonable modeling capacity. In contrast, removing temporal encoding causes the most severe performance deterioration on Wikipedia (dropping AP to nearly 80\%) which highlights its necessity for distinguishing fine-grained timestamp patterns in high-resolution temporal graphs.

In the inductive setting, performance patterns remain consistent. The removal of temporal encoding again has the most pronounced negative effect on Wikipedia. This suggests that in inductive scenarios where generalization to unseen nodes is required and temporal encoding plays an even greater role in preserving event chronology. The impact of removing $\mathbf{S}$ and decay mechanisms is also evident, though somewhat less dramatic than in the transductive setting, especially on Reddit. This may be due to Reddit’s lower temporal granularity and more regular interaction patterns.

\section{Concerns and Limitations}
\label{appendix:lims}
The proposed \our{} architecture demonstrated significant advancements, but several limitations warrant further investigation: 

The current design of \ours{} focuses on aggregating information from neighboring nodes, limiting its ability to capture high-order dependencies. This limitation may hinder the model's performance in scenarios requiring a deeper understanding of graph structures or relationships spanning multiple hops.

Although the \our{} effectively retains substantial information in its recurrent state, the fixed dimensionality of this state presents a scalability challenge. For super-large-scale and ultra-long-term dynamic graphs, increasing the dimensionality may be necessary to alleviate over-squashing. However, such modifications, could introduce significant computational overhead, necessitating careful tradeoffs between efficiency and expressiveness.

\end{document}